\documentclass[review,onecolumn,12pt,3p,authoryear]{elsarticle}

\usepackage{multirow}
\usepackage{subcaption}
\usepackage{graphicx}
\usepackage{float}
\usepackage{url}
\usepackage{booktabs} 
\usepackage{caption}  
\usepackage{tabularx} 
\usepackage{afterpage}
\usepackage{longtable}
\usepackage{amsmath}
\usepackage{amsfonts}
\usepackage{lineno,hyperref}

\makeatletter

\renewcommand{\paragraph}{\@startsection{paragraph}{4}{\z@}
  {3.25ex plus 1ex minus .2ex}  
  {0.1em}                         
  {\itshape}%
}
\makeatother

\begin{document}
	\sloppy
	\begin{frontmatter}
    
    \title{An Explainable Nature-Inspired Framework for Monkeypox Diagnosis: Xception Features Combined with NGBoost and African Vultures Optimization Algorithm}

\author[sem]{Ahmadreza Shateri}
\author[sem]{Negar Nourani}
\author[sem]{Morteza Dorrigiv}
\author[sem,lan]{Hamid Nasiri\corref{cor1}}
\cortext[cor1]{Corresponding author}

\address[sem]{Electrical and Computer Engineering Department, Semnan University, Semnan, Iran}
\address[lan]{School of Computing and Communications, Lancaster University, Lancaster, UK}

\ead{h.nasiri@lancaster.ac.uk}


\begin{abstract}
The recent global spread of monkeypox, particularly in regions where it has not historically been prevalent, has raised significant public health concerns. Early and accurate diagnosis is critical for effective disease management and control. In response, this study proposes a novel deep learning-based framework for the automated detection of monkeypox from skin lesion images, leveraging the power of transfer learning, dimensionality reduction, and advanced machine learning techniques. 
We utilize the newly developed Monkeypox Skin Lesion Dataset (MSLD), which includes images of monkeypox, chickenpox, and measles, to train and evaluate our models. The proposed framework employs the Xception architecture for deep feature extraction, followed by Principal Component Analysis (PCA) for dimensionality reduction, and the Natural Gradient Boosting (NGBoost) algorithm for classification. To optimize the model's performance and generalization, we introduce the African Vultures Optimization Algorithm (AVOA) for hyperparameter tuning, ensuring efficient exploration of the parameter space. 
Our results demonstrate that the proposed AVOA-NGBoost model achieves state-of-the-art performance, with an accuracy of 97.53\%, F1-score of 97.72\% and an AUC of 97.47\%. Additionally, we enhance model interpretability using Grad-CAM and LIME techniques, providing insights into the decision-making process and highlighting key features influencing classification. This framework offers a highly precise and efficient diagnostic tool, potentially aiding healthcare providers in early detection and diagnosis, particularly in resource-constrained environments.
\end{abstract}


\end{frontmatter}

\section{Introduction}

Since May 2022, monkeypox cases have emerged in countries where the disease had not previously been observed. As of the writing of this paper, 95,196 confirmed cases have been reported in 115 non-endemic countries, and 4,322 cases in 7 endemic countries \citep{CDC2022Monkeypox}. According to the World Health Organization (WHO) report, this is the first time that monkeypox cases have been simultaneously reported from both endemic and non-endemic countries in different geographical regions. While the COVID-19 pandemic continues to challenge the world, the emergence of a new outbreak caused by the monkeypox virus has raised concerns among public health authorities about whether it could constitute a new threat \citep{Yang2022}.

The causative agent of monkeypox, known as Monkeypox Virus (MPXV), like variola, cowpox, and vaccinia viruses, belongs to the orthopoxvirus genus and is a member of the Poxviridae family, which can lead to a wide range of diseases \citep{Realegeno2017}. Although the exact mode of MPXV transmission to humans remains largely unknown, most infections are presumed to be zoonotic, transmitted from animals to humans via direct or indirect contact. In some cases, human-to-human transmission occurs, likely through large respiratory droplets or via direct or indirect contact with body fluids, lesion material, contaminated fomites, or other substances \citep{Alakunle2020}. The disease was first detected in 1958 when two pox-like cases were observed in colony-maintained monkeys for research purposes. Subsequently, the first human case of the disease was reported in 1970 in the Democratic Republic of the Congo \citep{Kumar2022}.

After exposure, symptoms may not appear for days or even weeks. Early signs of monkeypox resemble influenza and include fever, chills, headaches, aches, muscle pains, fatigue, and swollen lymph nodes \citep{Fatima2022}. A rash typically develops a few days later, initially manifesting as painful, red, flat papules. These raised areas eventually progress into fluid-filled blisters that harden and fall off within two to four weeks. It is important to note that not all patients exhibit every symptom \citep{Adalja2022}. 
The clinical manifestations of monkeypox include symptoms and lesions that are often difficult to distinguish from smallpox \citep{Rizk2022}. While monkeypox typically presents with milder symptoms compared to the historically more severe smallpox, it can still be fatal, with a mortality rate of 1\% to 10\%, which is higher in children and young adults, and the course is more severe in individuals with immunodeficiency. Lymphadenopathy, or swelling of the lymph nodes, is observed in over 90\% of patients and serves as a distinguishing feature from smallpox. Owing to the nonspecific nature of its symptoms and lesions, a wide range of differential diagnoses must be considered, including chickenpox, molluscum contagiosum, measles, rickettsial infections, bacterial skin infections (such as those caused by Staphylococcus aureus), anthrax, scabies, syphilis, and drug reactions, among other noninfectious causes of rash \citep{Petersen2019}.

Clinical distinction among rash illnesses is challenging in the absence of diagnostic tests. These assessments play a crucial role in identifying monkeypox and can classify the disease using clinical specimens. These tests are most potent when they are combined with patient clinical information such as vaccination history \citep{Koenig2022}. Common methods, such as viral isolation from clinical specimens, electron microscopy, and immunohistochemistry, remain valid but require advanced technical skills and training. Specimens can be analyzed using real-time polymerase chain reaction (PCR) to assess the presence of orthopoxvirus or MPXV in a lesion sample.  Although PCR is effective for detecting viral DNA, its use is typically confined to major laboratories, making it less practical in rural or resource-poor settings.

The Tetracore Orthopox BioThreat Alert offers an alternative that does not require advanced equipment or specialized technicians and is thus suitable for low-resource areas. However, because this test is designed to diagnose orthopoxvirus species in general rather than specifically monkeypox, its use is ideally limited to regions where the disease is endemic \citep{MacNeil2009}. Considering these potentials, a significant conclusion was reached: A novel approach is required to address this gap.

Recent advancements in areas such as artificial intelligence and machine learning have made them one of the most useful tools for clinicians \citep{Bohr2020}. Deep learning is a subfield of artificial intelligence that aids in creating a model, autonomously extracting the features without the need for human participation, training the model, and generating the results \citep{Myszczynska2020}. Imaging techniques of all types are already in use in the medical field, assisting medical professionals in making diagnoses of a wide variety of diseases, including breast cancer \citep{Abbasniya2022} and other respiratory conditions such as pneumonia and tuberculosis \citep{Haloi2018}, as well as COVID and other conditions. The analysis of medical images via deep learning has been extensively studied recently. There is a need in this area due to the large number of monkeypox cases and the deficiency of testing kits. Due to the scarcity of specialist clinicians, it has been difficult to provide one to every hospital. In addition, the deep learning model can assist in addressing challenges such as the limited availability of RT-PCR kits, high costs, and lengthy wait times for diagnostic results, potentially complementing traditional testing methods when specialist clinicians are unavailable \citep{Ozturk2020}. Deep learning strategies have also been examined to determine whether machine learning could provide a viable solution to the problem of developing an effective triage strategy for the diagnosis of monkeypox disease. In this research, a deep learning model is deployed to enhance the accuracy of the diagnosis of monkeypox.

This analysis focuses on the performance of monkeypox detection using a newly developed monkeypox image dataset called the Monkeypox Skin Lesion Dataset (MSLD). To address the challenges in monkeypox diagnosis, we employed deep learning-based convolutional neural network (CNN) models and boosting algorithms. CNNs, widely recognized for their ability to learn features independently, have become the most popular approach for medical imaging classification. Specifically, we utilized the Xception architecture, a state-of-the-art technology known for reducing vanishing gradient issues, enhancing feature reuse, and minimizing the number of parameters. Xception's design incorporates linear residual connections, which improve gradient flow during training and enable effective propagation of information across layers. These residual connections link layers within each middle flow module rather than linking all layers. Compared to other image processing technologies, Xception has a more complex architecture due to its depthwise separable convolutions, allowing it to capture more visual information, making it particularly suitable for this task \citep{Chollet2016}.
The main goal of this study is to develop a dependable framework for detecting patients with monkeypox by analyzing skin lesion images. We propose an approach with high precision to improve the speed and accuracy of detection in order to distinguish from non-monkeypox (chickenpox and measles). To validate the proposed method, a series of experiments were conducted, and the results were compared with other competing feature extraction and parameter optimization algorithms. Statistical tests further confirmed the stability and effectiveness of our approach. The contributions of this work can be summarized as follows:

\begin{enumerate}
    \item We investigate potentials for early diagnosis and screening of monkeypox from skin lesion images utilizing deep learning-based models, including ResNet50V2, ResNet101V2, ResNet152V2, InceptionV3, InceptionResNetV2, MobileNet, MobileNetV2, DenseNet121, DenseNet169, DenseNet201, NASNetMobile, NASNetLarge, and Xception architectures.
    
    \item We implement the NGBoost model, a relatively new probabilistic approach, for monkeypox classification. The model leverages its probabilistic prediction capabilities and ability to handle complex, high-dimensional data while providing well-calibrated uncertainty estimates.
    
    \item The African Vultures Optimization Algorithm (AVOA), a novel nature-inspired metaheuristic optimization technique, is implemented to automatically tune the hyperparameters of the NGBoost model. This optimization strategy enhances the model's performance by efficiently exploring the hyperparameter space and identifying optimal configurations, resulting in improved classification accuracy and generalization capabilities.
    
    \item We enhance model interpretability through Grad-Cam and LIME techniques, providing valuable insights into the decision-making process and identifying key features that influence the classification of monkeypox lesions.
    
    \item This study addresses the challenge of early monkeypox detection by introducing an improved and efficient approach based on transfer learning, which is valuable for enhancing disease surveillance and control, especially for a rare disease like monkeypox.
\end{enumerate}

This remaining paper has been organized in the following way:

Section \ref{sec:rws} reviews related works on deep learning models in medical image analysis. Section \ref{sec:dd} details the Monkeypox Skin Lesion Dataset (MSLD) and data augmentation techniques. Section \ref{sec:metd} explains the methodology, including data preprocessing and feature extraction using pre-trained CNNs. It also outlines the use of classifiers such as LightGBM, SVM, XGBoost, and NGBoost, along with hyperparameter optimization using the African Vulture Optimization Algorithm (AVOA). Section \ref{sec:exp_res} presents the evaluation of models based on accuracy, F1-Score, and AUC, and discusses model interpretability through Grad-CAM. Finally, section \ref{sec:conc} concludes with the results and future research directions.

\section{Related Works}
\label{sec:rws} 
In this section, we revisit some of the recent methods related to deep learning with its applications to medical image analysis.

Medical image analysis using deep learning techniques has experienced a significant surge in recent years due to the widespread availability of sophisticated hardware. Typically, for a classification-based problem, more emphasis has been placed on feature extraction, which is arguably the most important aspect of any representation learning task.

An ensemble learning framework for detecting the monkeypox virus from skin lesion images was developed by \citep{Pramanik2023}. Their approach involved fine-tuning three pre-trained deep learning models—Inception V3, Xception, and DenseNet169—on the MSLD. The researchers extracted prediction probabilities from these base learners and incorporated them into an ensemble framework. They proposed a Beta function normalization scheme to efficiently aggregate the complementary information from the base learners, followed by a sum rule ensemble to combine the outcomes. The framework was evaluated on a public monkeypox lesion dataset using five-fold cross-validation. It achieved strong performance, with 93.39\% accuracy, 88.91\% precision, 96.78\% recall and 92.35\% F1 score.

\citep{Almufareh2023} presented a non-invasive, non-contact computer vision methodology for monkeypox diagnosis by analyzing skin lesion images. They employed deep learning techniques to classify lesions as either monkeypox positive or negative. The proposed approach was evaluated on two public datasets – Kaggle Monkeypox Skin Image Dataset (MSID) and MSLD. Multiple deep learning models were benchmarked using sensitivity, specificity, and balanced accuracy metrics. The proposed methodology yielded promising results, with MobileNet achieving 96.55\% balanced accuracy, 0.93 specificity, and maximum sensitivity on the MSID dataset. In contrast, InceptionV3 demonstrated the best metrics on the MSLD dataset with 94\% balanced accuracy, maximum specificity, and 0.88 sensitivity.

The comparative analysis of multiple pre-trained CNN models was the focus of research by \citep{Saha2023}. Their study utilized various architectures including VGG-16, VGG-19, Restnet50, Inception-V3, Densnet, Xception, MobileNetV2, Alexnet, Lenet, and majority Voting for classification. They combined multiple datasets such as monkeypox vs. chickenpox, monkeypox vs. measles, monkeypox vs. normal, and monkeypox vs. all diseases. In their experiments, majority voting achieved 97\% accuracy in identifying monkeypox versus chickenpox, Xception reached 79\% accuracy for monkeypox versus measles, MobileNetV2 obtained 96\% accuracy for monkeypox versus normal skin, and LeNet performed at 80\% accuracy for monkeypox versus all diseases. Their goal was to compare different CNN architectures for classifying monkeypox against other similar diseases and normal skin.

To address the challenge of distinguishing between monkeypox and measles symptoms, \citep{Ariansyah2023} developed an image classification approach using CNN architecture and VGG-16 transfer learning. They achieved high accuracy of 83.333\% at epoch 15 in differentiating monkeypox and measles symptoms. Their goal was to develop an automated image classifier to distinguish between these two diseases based on visual presentations.

\citep{Sitaula2022} compared 13 different pre-trained deep learning models for detecting the monkeypox virus. They fine-tuned all the models by adding universal custom layers and analyzed the results using precision, recall, F1-score, and accuracy. After identifying the best performing models, they ensembled them using majority voting on the probabilistic outputs to improve overall performance. Experiments were done on a public dataset, achieving 85.44\% precision, 85.47\% recall, 85.40\% F1-score and 87.13\% accuracy with their proposed ensemble approach. This demonstrates the power of ensembling multiple deep learning models to enhance monkeypox detection capabilities.

The performance of three pre-trained CNN models—MobileNetV2, VGG16, and VGG19—was investigated by \citep{Irmak2022} using the open-source 2022 Monkeypox Skin Image Dataset. Among these architectures, MobileNetV2 emerged as the top performer with 91.38\% accuracy, 90.5\% precision, 86.75\% recall, and 88.25\% F1 score, while VGG16 and VGG19 achieved 83.62\% and 78.45\% accuracy respectively.

Taking a practical approach to monkeypox diagnosis, \citep{Sahin2022} developed a mobile app using a deep learning model for preliminary monkeypox diagnosis from camera images. The CNN model was trained on public lesion images using transfer learning. After model selection and conversion to TensorFlow Lite, they tested the app on devices with fast inference times. The system allows infected individuals to get a preliminary diagnosis before seeking expert confirmation. Tests showed 91.11\% accuracy.\\
\citep{Islam2022} introduced the Monkeypox Skin Image Dataset 2022, the largest public dataset of its kind so far. They utilized this dataset to study deep learning models for monkeypox detection from skin images. Their study found state-of-the-art deep AI models have great potential for detecting monkeypox, achieving 83\% accuracy.

The integration of Convolutional Block Attention Modules (CBAM) with deep transfer learning was explored by \citep{Haque2022} for image-based monkeypox disease classification. They implemented five deep learning models - VGG19, Xception, DenseNet121, EfficientNetB3, and MobileNetV2-integrated with channel and spatial attention. Comparative analysis showed an Xception-CBAM model with Dense layers performed best, achieving 83.89\% validation accuracy in classifying monkeypox versus other diseases.

Unlike previous studies that utilized visual datasets and focused on feature extraction techniques from dermatological images to diagnose monkeypox, \citep{Farzipour2023} created a new purely textual dataset based on data collected and published by Global Health and used by the WHO to show the relationship between symptoms and monkeypox disease. Using this dataset, they analyzed and compared several machine learning methods, including Extreme Gradient Boosting (XGBoost), CatBoost, LightGBM, Support Vector Machine (SVM), and Random Forest. Their goal was to provide an ML model for diagnosing monkeypox based on symptoms. In their experiments, XGBoost achieved the best performance. To evaluate model robustness, they used k-fold cross-validation, reaching an average accuracy of 90\% across five different test set splits. Additionally, they utilized Shapley Additive Explanations (SHAP) to examine and explain the outputs of the XGBoost model.

Despite significant progress in using deep learning for monkeypox detection, existing methods face key limitations. Many approaches depend on ensemble techniques or pre-trained architectures without optimizing hyperparameters effectively, which can hinder performance. Furthermore, while interpretability techniques are sometimes utilized, their integration into workflows remains inconsistent, reducing the practical trustworthiness of model predictions.

Our proposed framework addresses these challenges by leveraging the AVOA for automated and efficient hyperparameter tuning. This results in a model that performs exceptionally well in terms of both accuracy and speed of prediction. Additionally, we incorporate interpretability techniques such as Grad-CAM and LIME to ensure a transparent and clinically relevant decision-making process. These enhancements enable a more robust and practical framework for monkeypox detection compared to prior studies.

\section{Dataset Description}
\label{sec:dd}
A publicly available dataset on Kaggle (an online community for data scientists and machine learning practitioners) served as the basis for the experiments conducted as part of this body of work.

\subsection{Data collection}
The dataset used to train the transfer learning models is titled the “Monkeypox Skin Lesion Dataset” (MSLD) \citep{Ali2022}. It is a binary classification dataset on Kaggle for monkeypox versus non-monkeypox images. The MSLD was generated by collecting and analyzing images obtained from various web-scraping sources, including news portals, websites, and public case reports.

\begin{figure}[htbp]
    \centering
    \begin{subfigure}[t]{0.13\textwidth} 
        \centering
        \includegraphics[width=\linewidth]{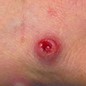}
        \caption{}
        \label{fig:sample1}
    \end{subfigure}
    \hfill
    \begin{subfigure}[t]{0.13\textwidth}
        \centering
        \includegraphics[width=\linewidth]{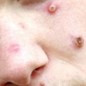}
        \caption{}
        \label{fig:sample2}
    \end{subfigure}
    \hfill
    \begin{subfigure}[t]{0.13\textwidth}
        \centering
        \includegraphics[width=\linewidth]{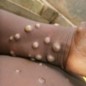}
        \caption{}
        \label{fig:sample3}
    \end{subfigure}
    \hfill
    \begin{subfigure}[t]{0.13\textwidth}
        \centering
        \includegraphics[width=\linewidth]{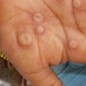}
        \caption{}
        \label{fig:sample4}
    \end{subfigure}
    \hfill
    \begin{subfigure}[t]{0.13\textwidth}
        \centering
        \includegraphics[width=\linewidth]{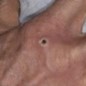}
        \caption{}
        \label{fig:sample5}
    \end{subfigure}
    \hfill
    \begin{subfigure}[t]{0.13\textwidth}
        \centering
        \includegraphics[width=\linewidth]{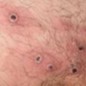}
        \caption{}
        \label{fig:sample6}
    \end{subfigure}
    \hfill
    \begin{subfigure}[t]{0.13\textwidth}
        \centering
        \includegraphics[width=\linewidth]{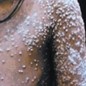}
        \caption{}
        \label{fig:sample7}
    \end{subfigure}
    
    \vspace{1em} 
    
    \begin{subfigure}[t]{0.13\textwidth}
        \centering
        \includegraphics[width=\linewidth]{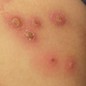}
        \caption{}
        \label{fig:sample8}
    \end{subfigure}
    \hfill
    \begin{subfigure}[t]{0.13\textwidth}
        \centering
        \includegraphics[width=\linewidth]{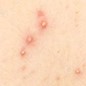}
        \caption{}
        \label{fig:sample9}
    \end{subfigure}
    \hfill
    \begin{subfigure}[t]{0.13\textwidth}
        \centering
        \includegraphics[width=\linewidth]{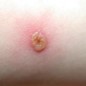}
        \caption{}
        \label{fig:sample10}
    \end{subfigure}
    \hfill
    \begin{subfigure}[t]{0.13\textwidth}
        \centering
        \includegraphics[width=\linewidth]{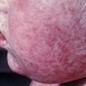}
        \caption{}
        \label{fig:sample11}
    \end{subfigure}
    \hfill
    \begin{subfigure}[t]{0.13\textwidth}
        \centering
        \includegraphics[width=\linewidth]{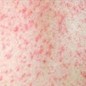}
        \caption{}
        \label{fig:sample12}
    \end{subfigure}
    \hfill
    \begin{subfigure}[t]{0.13\textwidth}
        \centering
        \includegraphics[width=\linewidth]{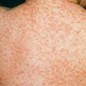}
        \caption{}
        \label{fig:sample13}
    \end{subfigure}
    \hfill
    \begin{subfigure}[t]{0.13\textwidth}
        \centering
        \includegraphics[width=\linewidth]{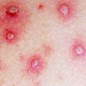}
        \caption{}
        \label{fig:sample14}
    \end{subfigure}

    \caption{Illustrative example of the dataset. Images a-g: Class 'MonkeyPox'. Images h–n: Class 'Others'.}
    \label{fig:sample_images}
\end{figure}

The dataset comprises 228 images, including 102 monkeypox cases and 126 non-monkeypox cases (chickenpox and measles). The images have a resolution of 224 × 224 × 3. \autoref{fig:sample_images} presents some examples of images from both classes.

\subsection{Data augmentation}

The dataset obtained included an original version that had been augmented by its creators. In training deep neural networks, dataset augmentation is used to increase the size of the dataset to enhance network performance \citep{Lewy2023}. The creators applied various augmentation techniques, including rotation, translation, reflection, shear, hue, saturation, contrast and luminance jitter, noise, and scaling. These methods aim to diversify the dataset and improve the model's ability to generalize \citep{Nayak2023}. By employing these methods, the model is exposed to a broader range of variations and acquires a deeper understanding of the images' underlying features.

\begin{figure}[htbp]
    \centering
    \begin{subfigure}[t]{0.13\textwidth} 
        \centering
        \includegraphics[width=\linewidth]{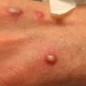}
        \caption{}
        \label{fig:aug1}
    \end{subfigure}
    \hfill
    \begin{subfigure}[t]{0.13\textwidth}
        \centering
        \includegraphics[width=\linewidth]{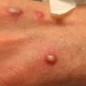}
        \caption{}
        \label{fig:aug2}
    \end{subfigure}
    \hfill
    \begin{subfigure}[t]{0.13\textwidth}
        \centering
        \includegraphics[width=\linewidth]{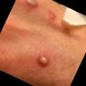}
        \caption{}
        \label{fig:aug3}
    \end{subfigure}
    \hfill
    \begin{subfigure}[t]{0.13\textwidth}
        \centering
        \includegraphics[width=\linewidth]{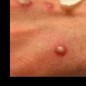}
        \caption{}
        \label{fig:aug4}
    \end{subfigure}
    \hfill
    \begin{subfigure}[t]{0.13\textwidth}
        \centering
        \includegraphics[width=\linewidth]{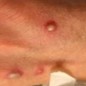}
        \caption{}
        \label{fig:aug5}
    \end{subfigure}
    \hfill
    \begin{subfigure}[t]{0.13\textwidth}
        \centering
        \includegraphics[width=\linewidth]{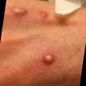}
        \caption{}
        \label{fig:aug6}
    \end{subfigure}
    \hfill
    \begin{subfigure}[t]{0.13\textwidth}
        \centering
        \includegraphics[width=\linewidth]{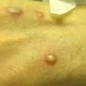}
        \caption{}
        \label{fig:aug7}
    \end{subfigure}
    
    \vspace{1em} 
    
    \begin{subfigure}[t]{0.13\textwidth}
        \centering
        \includegraphics[width=\linewidth]{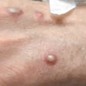}
        \caption{}
        \label{fig:aug8}
    \end{subfigure}
    \hfill
    \begin{subfigure}[t]{0.13\textwidth}
        \centering
        \includegraphics[width=\linewidth]{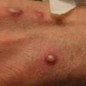}
        \caption{}
        \label{fig:aug9}
    \end{subfigure}
    \hfill
    \begin{subfigure}[t]{0.13\textwidth}
        \centering
        \includegraphics[width=\linewidth]{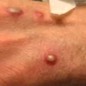}
        \caption{}
        \label{fig:aug10}
    \end{subfigure}
    \hfill
    \begin{subfigure}[t]{0.13\textwidth}
        \centering
        \includegraphics[width=\linewidth]{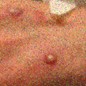}
        \caption{}
        \label{fig:aug11}
    \end{subfigure}
    \hfill
    \begin{subfigure}[t]{0.13\textwidth}
        \centering
        \includegraphics[width=\linewidth]{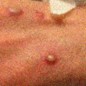}
        \caption{}
        \label{fig:aug12}
    \end{subfigure}
    \hfill
    \begin{subfigure}[t]{0.13\textwidth}
        \centering
        \includegraphics[width=\linewidth]{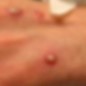}
        \caption{}
        \label{fig:aug13}
    \end{subfigure}
    \hfill
    \begin{subfigure}[t]{0.13\textwidth}
        \centering
        \includegraphics[width=\linewidth]{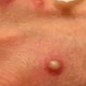}
        \caption{}
        \label{fig:aug14}
    \end{subfigure}
    
    \caption{Illustrative example of the fourteen-fold augmentation on an image from the 'Monkeypox' class. Image a: original image. Images b–n: Augmented images.}
    \label{fig:aug_images}
\end{figure}

In total, fourteen-fold augmentation was performed on the original dataset, bringing the number of images in each class to 1,428 in the 'Monkeypox' class and 1,764 in the 'Others' class. \autoref{fig:aug_images} shows some of the augmented images, which were stored in the augmented images folder.

The dataset includes both original and augmented images (Train, Test, Validation) with a proportion of 70:10:20. We used the data from the augmented Train folder for the training and validation phases, which contains 2,142 images. A detailed distribution of the dataset is provided in \autoref{tab:data_dist}.


\begin{table}[h]
	\centering
    \caption{Distribution of MSLD}
    \label{tab:data_dist}
	\footnotesize
    \begin{tabular}{@{\extracolsep{\fill}}lccc}
    \toprule
    Class Label & No. of Original Images & No. of Unique Patients & No. of Augmented Images \\ 
    \midrule
    Monkeypox & 102 & 55 & 1428 \\
    Others    & 126 & 107 & 1764 \\
    \midrule
    Total     & 228 & 162 & 3192 \\ 
    \bottomrule
	\end{tabular}
\end{table}

\section{Methodology}
\label{sec:metd}

The main idea behind the proposed model for classifying monkeypox and non-monkeypox cases is transfer learning, initially introduced by Stevo Bozinovski and Ante Fulgosi for use in neural networks \citep{Bozinovski2020}.

The process began by loading the monkeypox image dataset into the working environment. Following this, data preprocessing procedures were executed, including encoding class labels using a label encoder and applying normalization to images \citep{Li2020}. As a result, less redundant data were provided as input to the network.

In the next step, for feature extraction, the images were fed into pre-trained feature extraction models implemented through the Keras open-source library. Deep feature transfer learning involves using a neural network trained on a large-scale dataset, such as ImageNet, which includes hundreds of classes, to predict labels for other datasets \citep{Deng2009}. The proposed method was implemented by removing the fully connected layers, which typically classify images based on convolutionally extracted features, and replacing them with an alternate classifier.

The extracted features were subsequently transferred to another classifier for training. This added classifier could either be a fully connected layer or an external classification algorithm.

The proposed methodology involved several key steps:
\begin{itemize}
\item Data Preprocessing
\item Model Development
\item Training and Testing the Model (using cross-validation)
\item Hyperparameter Settings
\end{itemize}

\subsection{Data preprocessing}

The data preprocessing step includes feature scaling and data resizing, as illustrated below:

The first step involved exporting the dataset to manipulate and prepare the data. This step ensured that the data could be effectively utilized by various machine learning models for accurate detection of monkeypox cases. Data preprocessing specifically involved operations such as resizing the image, executing feature scaling, and extracting labels from images.

Labels were first extracted from the images using OpenCV methods, after which the images were downsized to 285×285 pixels. Following this, the extracted labels were encoded to make them compatible with classifiers, and the images were converted into NumPy arrays for subsequent use. In this process, scaling the features was essential to improve the models' performance.

Feature scaling, particularly normalization, standardizes input data to a specific range by processing the independent variables. Various normalization techniques exist, including Min-Max, Z-score, and Decimal scaling. This study employed Min-Max normalization, a method that preserves the relationships within the original data through linear transformation. The normalization process involved dividing each pixel value by the maximum possible value (255) to scale the data into the range [0,1] \citep{Pei1995}.

It follows that the normalized value is denoted by the symbol $x^{\prime}$.
\begin{equation}
x^{\prime}=\frac{x-\min (x)}{\max (x)-\min (x)}
\end{equation}
where $x$ is the original intensity of the image \citep{Bala2023}.

\subsection{Model development}

The model development stage includes pre-trained deep learning models, a dimensionality reduction method, some machine learning classifiers, and the proposed model.

\subsubsection{Deep learning model}

Extracting features is a process of transforming the initial data into usable information. The main objective of feature extraction is to reduce the quantity of data that must be processed while effectively characterizing the initial data. In image processing, feature extractors are beneficial for discovering properties such as shapes, edges, and movements. Feature extraction is advantageous when fewer resources are required for processing without losing key or essential data. The model or a subset of the model can preprocess the input to generate an output (i.e., a vector) for each input image, which can subsequently be used as training data for a new model \citep{Khalid2014}.

In this study, thirteen types of deep learning pre-trained models were employed for feature extraction: Xception, ResNet50V2, ResNet101V2, ResNet152V2, InceptionV3, InceptionResNetV2, MobileNet, MobileNetV2, DenseNet121, DenseNet169, DenseNet201, NASNetMobile, and NASNetLarge. Two essential parameters were configured during this process: the pooling type (whether to include or exclude the last layer) and the input shape, which was standardized to 285 × 285 for all models. A concise overview of the Xception model is presented below.

 \paragraph{Xception}
Xception is a deep convolutional neural network architecture proposed by \citep{Chollet2016} as a "depthwise separable variant of Inception". In this architecture, the n×n spatial convolution applied channel-wise is referred to as "depthwise convolution".

The central idea behind Xception is the hypothesis that cross-channel correlations and spatial correlations in convolutional neural networks can be entirely decoupled. To explore this hypothesis, Xception employs depthwise separable convolutions, which involve two key steps: first, depthwise spatial convolution is applied to each input channel independently, and then pointwise convolution projects the output channels onto a new channel space.

The Xception architecture comprises 71 layers in total, with 36 convolutional layers organized into 14 modules. The initial and final modules are designed to gradually adjust dimensionality, while the 12 intermediate modules form the core feature extraction "middle flow."

Specifically, the architecture begins with an "entry flow" consisting of three modules, which utilize depthwise separable convolutional layers to extract features and reduce spatial resolution. This is followed by eight repeating middle flow modules, each equipped with linear residual connections surrounding the depthwise separable convolutions to ensure efficient information propagation. These middle flow modules progressively extract increasingly complex features. Finally, the architecture concludes with an "exit flow" module, incorporating separable convolutions and global average pooling for classification.

Among the 14 modules, all but the first and last include residual connections. These linear residual connections encase the depthwise separable convolutions within each middle flow module, enhancing gradient flow during training.

Altogether, the 36 convolutional layers across the 14 modules account for 22.9 million learnable parameters. When evaluated on ImageNet—a dataset comprising over 14 million images across 1,000 classes—Xception achieved a top-1 accuracy of 79\% and a top-5 accuracy of 94.5\%. This performance slightly surpasses that of Inception V3, which has a comparable parameter count but was specifically optimized for ImageNet classification. The improved results of Xception suggest that its decoupling of cross-channel and spatial correlations enables more efficient parameter utilization \citep{Chollet2016}.

\subsubsection{Machine learning classifiers}

Four different machine learning classifiers were employed in various model structures in this study to classify monkeypox disease based on deep features extracted by the deep neural network. The classifiers are briefly described below.

\paragraph{LightGBM}

The Gradient Boosting Decision Tree (GBDT) is a reliable machine learning model that iteratively trains weak classifiers (Decision Tree) to optimize performance and minimize overfitting \citep{Zhou2019}. In 2017, researchers from Microsoft and Peking University introduced LightGBM to address efficiency and scalability challenges observed in existing methods like GBDT and XGBoost, especially for large datasets and high-dimensional features \citep{Li2023}.

\begin{figure}[htbp]\centering
\includegraphics[width=0.5\linewidth]{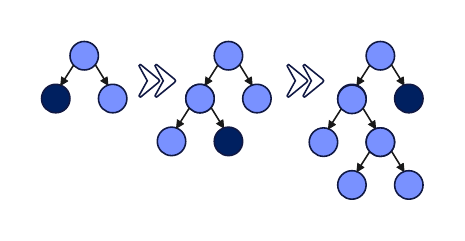}
\caption{LightGBM leaf-wise tree growth \citep{Abbasniya2022}.}
\label{fig:lgbm}
\end{figure}

LightGBM is a time-efficient, distributed lifting framework based on GBDT that offers faster training speed and lower memory occupation \citep{Sharma2020}. It employs a leaf-wise tree growth strategy (see \autoref{fig:lgbm}) which focuses on splitting leaves with higher gradients, in contrast to the level-wise growth strategy that treats all leaves at the same level equally in other GBDT tools. This innovative approach minimizes unnecessary computations and allows the generation of more complex trees \citep{Fu2023,Li_2022}.

Additionally, LightGBM incorporates advanced techniques such as gradient-based one-side sampling (GOSS) and exclusive feature bundling (EFB). GOSS reduces the number of samples by focusing on those with larger gradients, which have a more significant impact on information gain. EFB further enhances efficiency by bundling mutually exclusive features, effectively handling high-dimensional data \citep{Li_2022}. These features, combined with a histogram-based algorithm for data processing, enable smooth and accurate data scanning, clustering, and classification within shorter times.

\paragraph{SVM}

The Support Vector Machine (SVM), introduced by Vapnik in 1995, is a supervised learning algorithm known for its robustness in handling sparse, noisy, and high-dimensional data \citep{Cortes1995,Furey2000}. This method is based on the VC dimension theory and the structural risk minimization principle, which focuses on minimizing actual risk rather than empirical error \citep{Jing2015}.

The core concept of SVM is to identify an optimal decision boundary, or hyperplane, that separates two classes of data while maximizing the margin between them \citep{Ahmadipour2022,Chiu2020}.

The core concept of SVM is to identify an optimal decision boundary, or hyperplane, that separates two classes of data while maximizing the margin between them. For a dataset $\left\{x_i, y_i \mid i=1,2, \ldots, n\right\}$, where $x_i \in R^d$ represents input samples and $y_i \in\{-1,1\}$ are the corresponding labels, the hyperplane is defined by $w^T x_i+b=0$, with $w$ as the weight vector and $b$ as the bias term. The optimization problem seeks to minimize $\left\{\frac{1}{2}\|w\|^2+C \sum_{i=1}^n \xi_i\right\}$, where $C$ is a penalty parameter, and $\xi_i$ is a slack variable measuring the distance of misclassified points from the hyperplane \citep{Chiu2020}.

SVM is initially designed for linearly separable problems but can also address non-linear cases using kernel methods. Kernels map data from a low-dimensional, non-linear space to a higher-dimensional feature space where linear separation is achievable. Common kernels include linear, polynomial, Gaussian radial basis (RBF), and sigmoid functions. The RBF kernel, employed in this study, is effective for many practical applications. It is defined as $K\left(x, x_i\right)=e^{\left(-\gamma\left\|x-x_i\right\|^2\right)}$, where $\gamma=\frac{1}{2 \sigma^2}$, and $\sigma$ is the kernel width parameter.

The following function is the decision function utilized here
\begin{equation}\label{eq:svm1}
   f(x)=\operatorname{sign}\left(\sum_{i=1}^n \lambda_i y_i K\left(x, x_i\right)+b\right) 
\end{equation}
The decision function for SVM uses the kernel to compute \autoref{eq:svm1}. $f(x)$ computes the value of $x$ in the kernel function $K$ of every support vector through the new query vector $x$. In the equation, $b$ represents the offset of the hyperplane along the normal vector and is obtained through SVM training. $\lambda_i$ denotes the Lagrangian multiplier subject to $0 \leq \lambda_i \leq C$ \citep{Chiu2020}. By incorporating kernels, SVM effectively resolves the "curse of dimensionality" and enables robust classification in complex feature spaces.

\paragraph{XGBoost}

Extreme Gradient Boosting (XGBoost), introduced by Chen and Guestrin in 2016, is a highly scalable variant of the gradient boosting algorithm \citep{Chen2016}. The most notable distinction between XGBoost and other boosting decision trees is its scalability. In addition, despite its fast-learning capabilities, which increases the risk of overfitting, XGBoost's regularization technique prevents overfitting effectively, setting it apart from other gradient-boosting algorithms \citep{Abbasniya2022}. These features have recently made XGBoost widely adopted by data scientists, and it has achieved outstanding results in machine learning competitions.

The key idea behind XGBoost is to build an additive model by sequentially adding new weak learners (decision trees) that maximize the improvement in an objective function \citep{Ghaheri2024}. At each iteration, a new tree is constructed to predict the residuals (errors) of the model's predictions from the previous iteration. This is akin to performing gradient descent in function space, where the gradient is approximated by the latest decision tree \citep{Patnaik2021}.

Details of XGBoost are described below.

A given sample set with $n$ samples and $m$ features can be expressed as $D=\left\{\left(x_i, y_i\right)\right\}\left(|D|=n, x_i \in\right.$ $\mathbb{R}^m, y_i \in \mathbb{R}^n$ ), where $x$ is the eigenvalue, and $y$ is the true value \citep{Jiao2021}. A tree boosting model output $\hat{y}_i$ with $K$ trees is defined as follows:
\begin{equation}\label{eq:xgb1}
    \hat{y}_i=\sum_{k=1}^K f_k\left(x_i\right), \quad f_k \in F
\end{equation}
where $F$ represents the space of decision trees, defined as:
\begin{equation}\label{eq:xgb2}
    F=\left\{f(x)=\omega_q(x)\right\}\left(q: \mathbb{R}^m \rightarrow T, \omega \in \mathbb{R}^T\right)
\end{equation}
Each $f(x)$ divides a tree into structure part $q$ and leaf weights part $\omega$. Here $T$ denotes the number of leaf nodes. The objective function to be minimized is composed of two terms: the training loss $l$, which measures the discrepancy between predictions $\hat{y}_i$ and true values $y_i$, and a regularization term $\Omega$, which penalizes the complexity of the model:
\begin{equation}\label{eq:xgb3}
    O=\sum_{i=1}^n l\left(y_i, \hat{y}_i\right)+\sum_{k=1}^K \Omega\left(f_k\right)
\end{equation}
The regularization term is expressed as:
\begin{equation}\label{eq:xgb4}
\Omega(f)=\gamma T+\frac{1}{2} \lambda\|\omega\|
\end{equation}
where $\gamma$ penalizes the number of leaves $(T)$, and $\lambda$ is the penalty coefficient for the leaf weight $\omega$, which is usually constant. During training, a new tree is added to fit the residuals of the previous round \citep{Jiao2021}. Therefore, when the model has $t$ trees, it is expressed as follows:
\begin{equation}\label{eq:xgb5}
\hat{y}^t=\hat{y}^{t-1}+f_t(x)
\end{equation}
Substituting \autoref{eq:xgb5} into the objective function \autoref{eq:xgb3} yields the function
\begin{equation}\label{eq:xgb6}
O^{(t)}=\sum_{i=1}^n l\left(y_i, \hat{y}_i^{t-1}+f_t\left(x_i\right)\right)+\Omega\left(f_t\right)
\end{equation}
XGBoost approximates \autoref{eq:xgb6} by utilizing the second order Taylor expansion. The first-order gradient $g_i$ and second-order gradient $h_i$ of the loss function are calculated for each instance, enabling precise updates to the model:
\begin{equation}\label{eq:xgb7}
O^{(t)} \approx \tilde{O}^{(t)}=\sum_{i=1}^n\left[l\left(y_i, \hat{y}_i^{t-1}\right)+g_i f_t\left(x_i\right)+\frac{1}{2} h_i f_t^2\left(x_i\right)\right]+\Omega\left(f_t\right)
\end{equation}
To determine the optimal structure of the tree, the algorithm evaluates the weights of each leaf. For leaf $j$, the optimal weight is computed as:
\begin{equation}\label{eq:xgb8}
\omega_j^*=-\frac{\sum_{i \in I_j} g_i}{\sum_{i \in I_j} h_i+\lambda}
\end{equation}
where $I_j$ represents the set of instances in leaf $j$. Substituting this into the scoring function, the final objective for the tree becomes:
\begin{equation}\label{eq:xgb9}
\tilde{O}^{(t)}=-\frac{1}{2} \sum_{j=1}^T \frac{\left(\sum_{i \in I_j} g_i\right)^2}{\sum_{i \in I_j} h_i+\lambda}+\gamma T
\end{equation}
Define \autoref{eq:xgb9} as a scoring function to evaluate the tree structure $q(x)$ and find the optimal tree structures for classification.

\paragraph{NGBoost}

In the realm of boosting algorithms, several existing methodologies have proven effective in enhancing predictive accuracy. However, there is a substantial gap in managing probabilistic predictions and quantifying uncertainty. Methods like AdaBoost, Gradient Boosting, and XGBoost typically focus on reducing prediction error and often overlook estimating the uncertainty of predictions. They are primarily capable of estimating the conditional mean of the response variable as a function of the predictors \citep{Duan2020}. This oversight can reduce the model's ability to quantify uncertainty and make accurate probabilistic predictions in cases where precise assessment of uncertainty is essential, such as in medical diagnostics or financial risk analysis. Consequently, this gap highlights the need for a sophisticated boosting technique that not only improves prediction accuracy but also focuses on providing well-calibrated probability estimates and credible uncertainty measures.

A suitable candidate for addressing this issue is an innovative method called Natural Gradient Boost (NGBoost). NGBoost can be employed to calculate statistical uncertainty through the gradient boosting technique by using probabilistic forecasts (including actual point predictions) \citep{Jabeur2022}. In addition, the algorithm's probabilistic forecasting feature makes it feasible to acquire probabilities about the whole possible outcomes of any given search area rather than determining one speciﬁc output as the prediction \citep{Kavzoglu2022}.

In fact, NGBoost transforms the boosting process into a probabilistic framework using the natural gradient to compensate for the probabilistic prediction limitations of existing gradient boosting methods \citep{Dutta2020}.

In point prediction settings, the object of interest is an estimate of a scalar function like $E[y \mid x]$, where $x$ is the feature vector and $y$ is the prediction target. Inversely, the NGBoost algorithm considers the probabilistic prediction with a conditional probability distribution function $P_\theta(y \mid x)$ which is generated through the forecasting the parameters of the target distribution $\theta$. NGBoost assumes that $P_\theta(y \mid x)$ is of a specified parametric form, then estimate the $p$ parameters $\theta \in \mathbb{R}^p$ of the distribution as functions (base learners) of $x$ under the proper scoring rules $S(\theta, y)$ and fitting natural gradient $\nabla_\theta$ \citep{Chen2022,Duan2020}.

\begin{figure}[htbp]\centering
\includegraphics[width=0.95\linewidth]{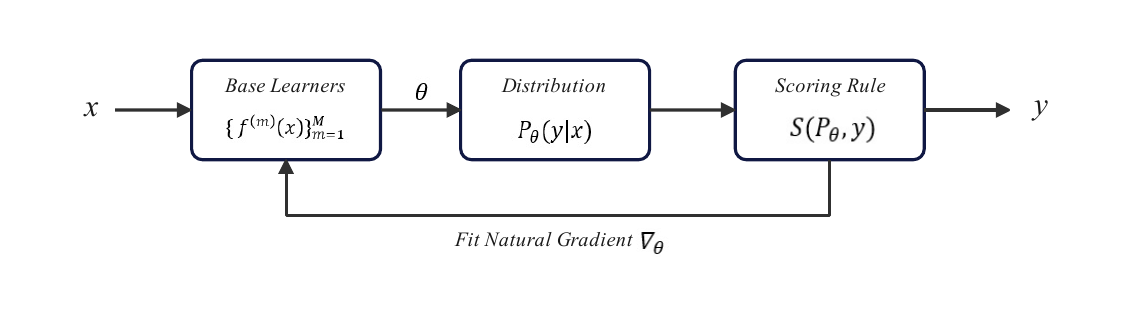}
\caption{NGBoost is modular with respect to choice of base learner, distribution, and scoring rule \citep{Duan2020}.}
\label{fig:ngb}
\end{figure}

The modular and scalable algorithm consists of three primary components: (1) The base learner (e.g., Decision Tree, etc.); (2) Parametric probability distribution (Bernoulli, Normal, Laplace, etc.); (3) The proper scoring rule (MLE, CRPS, etc.) \citep{Peng2020}. A conceptual representation of the NGBoost model is shown in \autoref{fig:ngb}.

In \autoref{fig:ngb}, $x_i$ represents the input features, $f^{(m)}$ is the base learner, $y_i$ is the target labels, and $\theta$ represents parametric probability density function and $M$ is the Boosting iterations \citep{Hussain2021}.

The working procedure of the NGBoost algorithm is given as follows:
\begin{enumerate}
    \item Compute the initial $\theta_0 \leftarrow \arg \min _\theta \sum_{i=1}^n S\left(\theta, y_i\right)$ Underline that $\theta$ can be a vector of parameters.
    \item Execute $M$ gradient boosting stages \citep{Kavzoglu2022}.
\end{enumerate}
In contrast to traditional boosting algorithms, the algorithm can be scaled to handle large variables or observations with reasonable complexity \citep{Jabeur2022}. Also, it should be noted that to the best of our knowledge, this is the first paper on the application of the NGBoost algorithm for diagnosing monkeypox.

NGBoost offers a flexible framework to customize the base learner, parametric probability distribution, and scoring rules as hyperparameters in advance before training \citep{Chen2022}. In order to obtain satisfactory results, except for these three components, we also tuned hyperparameters based on the training set using a meta-heuristic algorithm, which efficiently explores the parameter space through a heuristic search process to find the optimal set of parameters.

\subsubsection{Hyperparameter optimization}

A hyperparameter is defined as a higher-level concept of the model, such as its complexity or learnability. Hyperparameters are parameters established before the learning process and not parameters obtained through training. Generally, in the process of machine learning, it is necessary to optimize these parameters and select a set of optimal hyperparameters for the learner to enhance learning performance and effect \citep{Dong2020}. However, hyperparameter tuning is subjective and relies on trial-and-error methods and empirical judgment \citep{Bergstra2011}. The hyperparameter optimization algorithm overcomes the dependence on manual search, experience, and trial and error \citep{Jiao2021}.

With the increasing sophistication of various non-linear problems, there is an imperative need to develop further metaheuristic optimization algorithms based on natural phenomena and biological behaviors to efficiently solve these complex search and optimization problems.

The optimization process for hyperparameters comprises four basic components: an estimator with its target function, a search space (configuration area), and a searching or optimization scheme used to locate hyperparameter configurations.

The main procedure for hyperparameter configurations is as follows:

\begin{enumerate}
    \item Choose an objective function and performance measurements.
    \item Choose the parameters to be tuned, review their types, and identify the best optimization strategy.
    \item Provide the essential workings for the ML model as a baseline model utilizing the default parameter setup or common values.
    \item Initialize optimization with a wide search area as the viable hyperparameter domain defined by manual testing skills.
    \item Narrow the search area based on well-tested regions and explore new search areas.
    \item Return the most successful configuration of the hyperparameter as the final solution.
\end{enumerate}
During the design phase of a model, the optimal hyperparameters for the templates can be determined by effectively searching the hyperparameter space with optimization algorithms. The process of providing better parameter optimization comprises critical elements such as the predictor's objectives and outcomes, the need for a space to apprehend the data and optimize the combination, and the application of various evaluation functions that enable comparison of the results \citep{Jiao2021}.

This section provides an overview of the metaheuristic utilized to hybridize machine learning models.

\paragraph{African vulture optimization algorithm (AVOA)}

A novel metaheuristic swarm-based optimization algorithm that draws inspiration from the environment, AVOA mimics the foraging and predation behaviors of a species of vulture that lives mainly in Africa. Furthermore, compared to other metaheuristic algorithms, this algorithm is more adaptive and has a lower computational complexity \citep{Abdollahzadeh2021}.

The AVOA algorithm principle can be summarized into four primary phases as follows:

\textbf{Phase one: the best vultures' group determination}

The AVOA population consists of $N$ vultures, where $N$ is adjustable according to the problem's requirements. Initially, the population is generated randomly, and fitness is evaluated for all individuals. In each iteration, the solutions are grouped as follows: the best-performing solutions form the first-best vulture group, while the second-best solutions form the second-best vulture group. The remaining vultures are distributed between these two groups using a Roulette wheel approach based on selection probabilities calculated by:
\begin{equation}\label{eq:avoa1}
p_i=\frac{F_i}{\sum_{i=1}^n F_i}
\end{equation}
where $F_i$ is the fitness value of the $i^{\text {th }}$ vulture. The best vultures in the population are then assigned as:
\begin{equation}\label{eq:avoa2}
R(i)= \begin{cases}BestVulture_1, & \text { if } p_i=L_1 \\ BestVulture_2, & \text { if } p_i=L_2\end{cases}
\end{equation}
Parameters $L_1$ and $L_2$ are predetermined, ranging from 0 to 1 , with $L_1+L_2=1$. Higher $L_1$ values increase intensification, while higher $L_2$ values enhance diversification.

\textbf{Phase two: starvation rate of vultures}

This phase determines the starvation rate of vultures. When overstuffed, vultures explore farther, but when hungry, they stay near others with food. This behavior is modeled as:
\begin{equation}\label{eq:avoa3}
F=\left(2 \cdot \operatorname{rand}_1+1\right) \cdot z \cdot\left(1-\frac{\operatorname { iteration }_i}{\text { max iterations }}\right)+t
\end{equation}
\autoref{eq:avoa3} computes the satiation rate, which has been used to transfer the AVOA from the exploration phase to the exploitation one.
\begin{equation}\label{eq:avoa4}
t=h\left(\sin ^w\left(\frac{\pi}{2} \times \frac{\operatorname { iteration }_i}{\text { max iterations }}\right)+\cos \left(\frac{\pi}{2} \times \frac{\operatorname { iteration }_i}{\text { max iterations }}\right)-1\right)
\end{equation}
where $iteration_i$ is the current iteration, $z$ is a random value between -1 and 1 , and $h$ is a random number between -2 and 2. If $z$ is positive, the vulture is satisfied; if negative, it is starved. The parameter $w$ influences the transition between exploration and exploitation phases. When $|F|>1$, vultures are in the exploration phase.

\textbf{Phase three: exploration}

In this phase, vultures with high visual ability search for food and predict dying prey. The exploration strategy depends on the parameter $P_1$ which controls the search approach. Two techniques are employed:
\begin{enumerate}
    \item Focused Search: Around the best vulture in the group, encouraging local exploration. This strategy enables a focused investigation close to the current top solution.
    \item Global Search: Across the entire search space within predefined boundaries.
\end{enumerate}

The vulture's position in the next iteration is updated using:
\begin{equation}\label{eq:avoa5}
   P(i+1)=R(i)-D(i) \cdot F \quad \text { if } P_1 \geq \operatorname{rand}_{P_1} 
\end{equation}
\begin{equation}\label{eq:avoa6}
P(i+1)=R(i)-F+\operatorname{rand}_2 \cdot\left((u b-l b) \cdot \operatorname{rand}_3+l b\right) \quad \text { if } P_1<\operatorname{rand}_{P_1}
\end{equation}
where $P_1$ is a pre-set search parameter between 0 and 1 to control the search strategy, $R(i)$ is the optimal vulture obtained by \autoref{eq:avoa2} at the current iteration, $D(i)=|X \cdot R(i)-P(i)|, X$ is a coefficient vector for random movement, $F$ is the rate of starvation obtained by using \autoref{eq:avoa3}, $P(i+1)$ is the position of the vulture in the next iteration, and $u b$ and $l b$ are the upper and lower bounds of the search space.

\textbf{Phase four: exploitation}

If $|F|<1$, the AVOA enters the exploitation phase, which has two stages, each involving two strategies. The selection of each strategy depends on $P_2$ and $P_3$, both ranging between 0 and 1.

Stage 1: Rotational Flight and Siege Fight

When $0.5 \leq|F|<1$, vultures perform two tactics: rotational flight and siege fight. The choice between them is made by generating a random number $\operatorname{ran} d_{P_2}$ and comparing it with $P_2$. The following equations represent these strategies:

\begin{itemize}
    \item Siege Fight:
\end{itemize}
\begin{equation}\label{eq:avoa7}
\begin{aligned}
& P(i+1)=D(i) \times\left(F+\operatorname{rand}_4\right)-d(t) \quad P_2 \geq \operatorname{rand}_{P_2} \\
& d(t)=R(i)-P(i)
\end{aligned}
\end{equation}

\begin{itemize}
    \item Rotational Flight:
\end{itemize}
\begin{equation}\label{eq:avoa8}
\begin{aligned}
& P(i+1)=R(i)-\left(S_1+S_2\right) \\
& S_1=R(i) \times\left(\frac{\operatorname{rand}_5 \times P(i)}{2 \pi}\right) \times \cos (P(i)) \quad P_2<\operatorname{rand}_{P_2} \\
& S_2=R(i) \times\left(\frac{\operatorname{rand}_6 \times P(i)}{2 \pi}\right) \times \sin (P(i))
\end{aligned}
\end{equation}
where $rand_4$, $rand_5$, and $rand_6$ are random numbers between 0 and 1 , which are used to increase the randomness. The value of $\cos (P(i))$ or $\sin (P(i))$ is an array vector of size $n \times 1$, where $n$ represents the number of generation units.

Stage 2: Aggressive Siege and Levy Flight

When $|F|<0.5$, the second exploitation stage begins, involving more aggressive siege tactics. A strategy is selected using $P_3$ as follows:
\begin{itemize}
    \item Aggressive Siege:
\end{itemize}
\begin{equation}\label{eq:avoa9}
\begin{aligned}
& P(i+1)=\frac{A_1+A_2}{2}, \quad P_3 \geq \operatorname{rand}_{P_3} \\
& A_1=\operatorname{BestVulture}_1(i)-\frac{\operatorname{BestVulture}_1(i) \times P(i)}{\operatorname{BestVulture}_1(i) \times P(i)^2} \times F \\
& A_2=\operatorname{BestVulture}_2(i)-\frac{\operatorname{BestVulture}_2(i) \times P(i)}{\operatorname{BestVulture}_2(i) \times P(i)^2} \times F
\end{aligned}
\end{equation}
\begin{itemize}
    \item Levy Flight Strategy:
\end{itemize}
\begin{equation}\label{eq:avoa10}
P(i+1)=R(i)-|d(t)| \times F \times \operatorname{Levy}(d), \quad P_3<\operatorname{rand}_{P_3}
\end{equation}

where $BestVulture_1(i)$ and $BestVulture_2(i)$ are the first- and second-best vultures in the current iteration, $P(i)$ is the vector of the current vulture's position. $d(t)$, $R(i)$ and $F$ are as defined before, and $\operatorname{Levy}(d)$ stands for Le'vy flight which is used to increase the efficiency of the AVOA, and the mathematical expression of Levy ' flight $(L F)$ is as follows:
\begin{equation}\label{eq:avoa11}
\begin{aligned}
& L F(x)=0.01 \times \frac{u \times \sigma}{|v|^{\frac{1}{\beta}}} \\
& \sigma=\left(\frac{\Gamma(1+\beta) \times \sin \left(\frac{\pi \beta}{2}\right)}{\Gamma(1+\beta 2) \times \beta \times 2\left(\frac{\beta-1}{2}\right)}\right)^{\frac{1}{\beta}}
\end{aligned}
\end{equation}
Where both $u$ and $v$ are random numbers between 0 and 1 , and $\beta$ denotes a fixed constant of 1.5 \citep{Zhou2023}. 

\subsection{Proposed model}

In this section, we present our proposed model for the classification of monkeypox cases from skin lesion images. Our model seamlessly integrates transfer learning, dimensionality reduction, and advanced machine learning techniques to achieve state-of-the-art results. The following steps delineate our proposed model:

In the initial stage of our model, we focused on preparing the dataset to ensure it was suitable for training. This involved two critical preprocessing steps: label encoding and normalization. Label encoding was employed to represent the classes numerically, ensuring compatibility with machine learning algorithms. Additionally, we applied image normalization, which scales pixel values to the range [0, 1]. These preprocessing techniques not only facilitated computational efficiency but also improved the model's ability to discern meaningful features within the data.

Our model's foundation lies in the utilization of the Xception deep neural network, initialized with pre-trained weights from the ImageNet dataset. We configured Xception with an input shape of (285 × 285) and excluded the top layers for our specific task. Crucially, we set the network's layers as non-trainable, preserving the pre-trained knowledge encoded in the network.

Following feature extraction with Xception, we obtained a high-dimensional feature representation of our skin lesion images. This step enabled us to extract 2048 features, capturing intricate patterns and details crucial for accurate classification.

To mitigate the risk of overfitting and enhance computational efficiency, we employed Principal Component Analysis (PCA) to reduce the feature space to a more manageable 530 features. PCA is a dimensionality reduction technique used to address the challenge of highly correlated features and the "curse of dimensionality," where the number of features often exceeds the number of instances in datasets. By transforming the original high-dimensional data into a smaller set of uncorrelated variables, PCA retained the most significant information while eliminating redundancy. This reduction enhanced computational efficiency, mitigated the risk of overfitting, and simplified the feature space, making it more manageable for the classifier to process. As part of our method, PCA helped focus on the most informative aspects of the data, improving model performance and ensuring faster predictions.

In the final stage of our proposed model, the NGBoost algorithm was used to perform the final classification of monkeypox and non-monkeypox cases. Our NGBoost model was fine-tuned using the African Vultures optimization algorithm, ensuring optimal performance.

\begin{figure}[htbp]\centering
\includegraphics[width=0.95\linewidth]{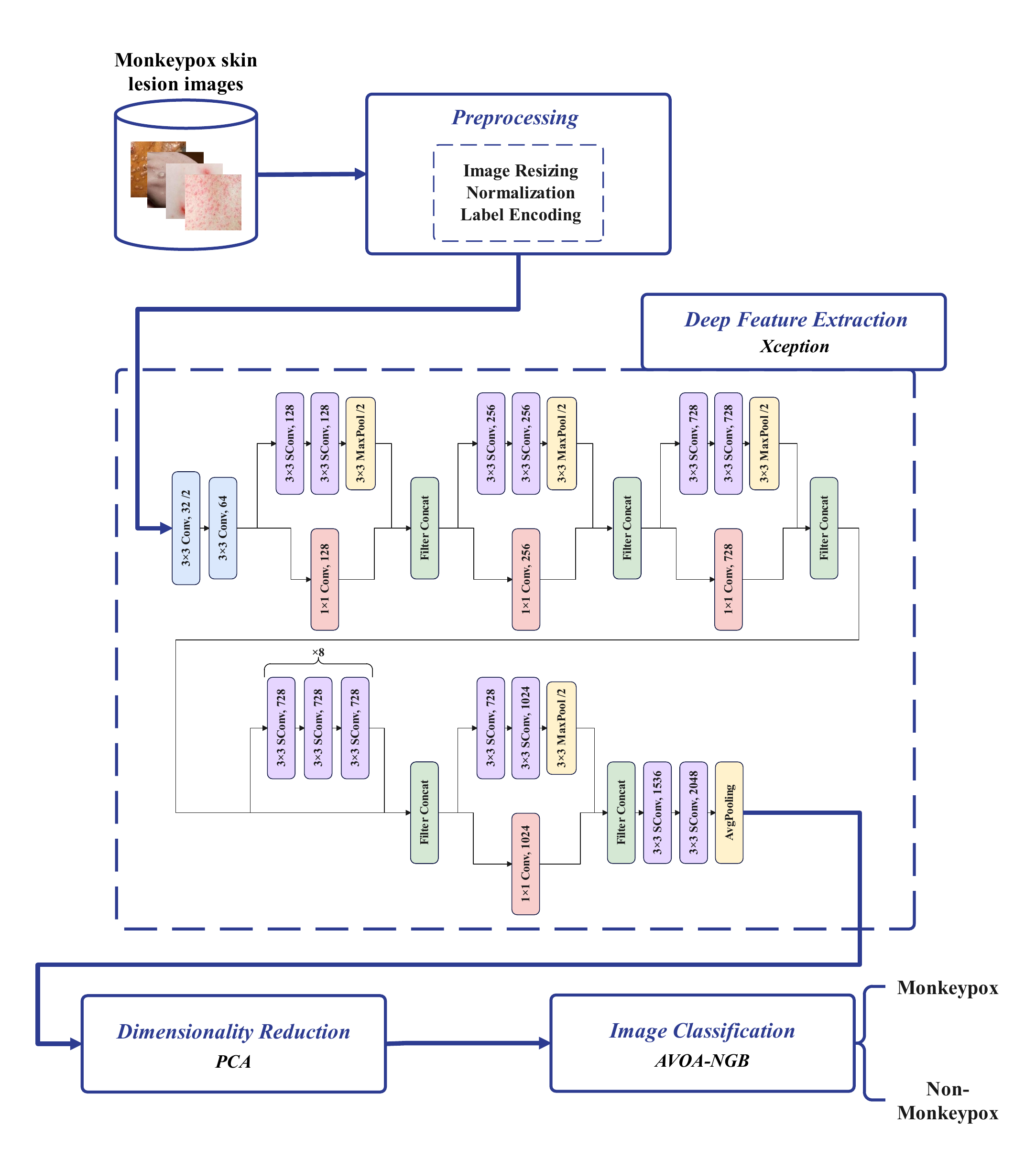}
\caption{Proposed methodological architecture.}
\label{fig:pm}
\end{figure}

\autoref{fig:pm} illustrates the steps of the proposed process and method for the classification and prediction of monkeypox disease.

\section{Evaluation metrics}
\label{em}

The overall experimental outcome is measured and presented using widely adopted statistical methods, including accuracy, precision, recall, specificity, and F1-score. These quantitative measurements have been selected based on their effectiveness for the classification task and their frequent use in closely related research.

The degree to which the model accurately classifies the images in the test dataset is referred to as its “accuracy.”

The term “precision” refers to the ability to precisely estimate the actual positive outcomes for both categories.

“Recall” represents the proportion of cases that are true positives in both categories. When false negatives are associated with a high cost, recall is the primary statistic used to select the optimal model.
“Specificity” identifies the number of correctly identified and calculated true negatives.

The “F1-score” is a metric that demonstrates the relationship between precision and recall. It is commonly used in machine learning and information retrieval to evaluate the performance of a classification model. The F1-score is computed as the harmonic mean of precision and recall, giving equal weight to both measures. A high F1-score indicates that the model achieves both high precision and high recall, meaning it can correctly identify the most relevant cases while minimizing false positives.

Performance comparisons can lead to misleading results if based solely on accuracy metrics. Such assessments must also consider the cost of errors. In this context, we evaluate the classifier's reliability using Cohen's kappa metric, which accounts for the extent to which agreement between actual and predicted class values can be explained by chance. Generally, kappa values range between -1 and 1. When the calculated kappa value for the classifier is 1, the classifier's performance is considered to be in perfect agreement with the true values.

In our dataset, monkeypox can be classified as true positive (\textit{TP}) or true negative (\textit{TN}) if individuals are correctly diagnosed, or as false positive (\textit{FP}) or false negative (\textit{FN}) if individuals are misdiagnosed. The formulas for the performance evaluation criteria used in this study are given below:
\begin{equation}
    \text { Accuracy }=\frac{T P+T N}{T N+T P+F P+F N}
\end{equation}
\begin{equation}
    \text { Precision }=\frac{T P}{T P+F P}
\end{equation}
\begin{equation}
    \text { Recall }=\frac{T P}{T P+F N}
\end{equation}
\begin{equation}
    \text { Specificity }=\frac{T N}{T N+F P}
\end{equation}
\begin{equation}
    F_1 \text {-score }=\frac{T P}{T P+\frac{1}{2}(F P+F N)}
\end{equation}
\begin{equation}
    \left\{\begin{array}{c}
    P_o=A C=\frac{T P+T N}{T P+T N+F P+F N} \\
    P_e=\frac{(T P+F P)(T P+F N)+(T N+F P)(T N+F N)}{(T P+T N+F P+F N)^2} \\
\kappa=\frac{P_o-P_e}{1-P_e}
\end{array}\right.
\end{equation}
With,  $P_o$ demonstrates total agreement probability and  $P_e$ agreement probability due to chance.

\section{Experimental Results}
\label{sec:exp_res}

In this section, we present and analyze the performance of our proposed method for monkeypox detection using the Monkeypox Skin Lesion Dataset (MSLD).

The cross-validation technique was adopted due to the small scale of the collected dataset, ensuring the efficacy of the proposed method. As part of the k-fold cross-validation strategy, the entire dataset was randomly divided into \textit{k} sub-folders of equal size. Subsequently, the learning function was trained using (\textit{k-1}) sub-folders of training data, while the remaining fold was used to test the model. Each individual data subset was utilized for either training or validation. After iterating through all data subsets in sequence, a cumulative average validation score was calculated. We employed the 5-fold cross-validation technique, where four folds (80\% of the data) were used for training, and the remaining fold (20\% of the data) was used to evaluate the model's test accuracy.

In the initial stage, 13 pre-trained deep learning models were used to extract features from the input images. These features were then supplied to four different machine learning classifiers: LightGBM, SVM, XGBoost, and NGBoost, to predict the final labels. The performance results of these models are presented in \autoref{tab:fe_cls_res}, where L, S, X, and N represent LightGBM, SVM, XGBoost, and NGBoost, respectively.

\afterpage{
\begin{clearpage}
\begin{scriptsize}
\begin{longtable}{@{\extracolsep{\fill}}llcccccc}
\caption{Classification results for different deep learning models.}
\label{tab:fe_cls_res}\\
\toprule
\multirow{2}{*}{Feature extractor} & \multirow{2}{*}{Classifier} & \multicolumn{6}{c}{Results} \\ 
\cmidrule(lr){3-8}
 & & Accuracy (\%) & Precision (\%) & Recall (\%) & Specificity (\%) & Kappa & F1-Score (\%) \\ 
\midrule
\endfirsthead 

\multicolumn{8}{c}{{\tablename\ \thetable{} -- Continuation of \autoref{tab:fe_cls_res}}} \\ \hline
\toprule
\multirow{2}{*}{Feature extractor} & \multirow{2}{*}{Classifier} & \multicolumn{6}{c}{Results} \\ 
\cmidrule(lr){3-8}
 & & Accuracy (\%) & Precision (\%) & Recall (\%) & Specificity (\%) & Kappa & F1-Score (\%) \\ 
\midrule
\endhead 

\midrule
\multicolumn{8}{r}{\raisebox{0.5ex}{\small Continued on next page}} \\[-1.5ex]
\midrule
\endfoot 
\bottomrule
\endlastfoot 
\multirow{4}{*}{Xception}    & L & 93.60          & 92.47 & 96.11 & 90.79 & 87.05 & 94.22          \\
                             & S & \textbf{95.89} & 94.75 & 97.88 & 93.63 & 91.68 & \textbf{96.27} \\
                             & X & 93.51          & 92.73 & 95.60 & 91.20 & 86.88 & 94.10          \\
                             & N & \textbf{95.15} & 93.65 & 97.70 & 92.20 & 90.16 & \textbf{95.61} \\ 
\midrule
\multirow{4}{*}{ResNet50V2}  & L & 85.48          & 87.78 & 85.16 & 85.87 & 70.77 & 86.43          \\
                             & S & 87.63          & 93.85 & 82.74 & 93.55 & 75.34 & 87.92          \\
                             & X & 84.83          & 86.99 & 84.75 & 85.03 & 69.48 & 85.83          \\
                             & N & 83.94          & 84.08 & 87.41 & 80.31 & 67.61 & 85.52          \\ 
\midrule
\multirow{4}{*}{ResNet101V2} & L & 94.91          & 94.26 & 96.49 & 93.11 & 89.71 & 95.35          \\
                             & S & 94.91          & 93.66 & 97.17 & 92.31 & 89.70 & 95.37          \\
                             & X & 94.68          & 94.08 & 96.24 & 92.96 & 89.25 & 95.12          \\
                             & N & 94.54          & 92.62 & 97.68 & 90.88 & 88.93 & 95.07          \\ 
\midrule
\multirow{4}{*}{ResNet152V2} & L & \textbf{95.15}  & 93.99 & 97.25 & 92.70 & 90.17 & 95.58          \\
                             & S & \textbf{95.33} & 94.09 & 97.51 & 92.78 & 90.54 & \textbf{95.76} \\
                             & X & 93.84          & 92.77 & 96.13 & 91.16 & 87.52 & 94.40          \\
                             & N & 93.93          & 91.45 & 97.93 & 89.23 & 87.68 & 94.57          \\ 
\midrule
\multirow{4}{*}{InceptionV3} & L & 83.75          & 86.73 & 82.79 & 85.00 & 67.38 & 84.68          \\
                             & S & 86.88          & 93.50 & 81.57 & 93.22 & 73.85 & 87.11          \\
                             & X & 83.33          & 86.05 & 82.84 & 84.14 & 66.54 & 84.35          \\
                             & N & 81.23          & 80.74 & 86.52 & 75.47 & 62.08 & 83.34          \\ 
\midrule
\multirow{4}{*}{InceptionResNetV2} & L & 84.03 & 86.32 & 83.98 & 84.16 & 67.86 & 85.10          \\
                             & S & 87.91          & 94.13 & 82.94 & 93.86 & 75.87 & 88.17          \\
                             & X & 84.45          & 86.93 & 84.07 & 84.90 & 68.71 & 85.46          \\
                             & N & 83.94          & 84.10 & 87.14 & 80.49 & 67.58 & 85.46          \\ 
\midrule
\multirow{4}{*}{MobileNet}   & L & 82.21          & 85.49 & 81.13 & 83.77 & 64.37 & 83.14          \\
                             & S & 84.36          & 91.72 & 78.38 & 91.73 & 68.94 & 84.41          \\
                             & X & 81.51          & 84.12 & 81.46 & 81.85 & 62.90 & 82.67          \\
                             & N & 79.83          & 80.04 & 84.62 & 74.75 & 59.32 & 82.00          \\ 
\midrule
\multirow{4}{*}{MobileNetV2} & L & 82.12          & 84.87 & 81.65 & 82.74 & 64.06 & 83.21          \\
                             & S & 83.61          & 88.81 & 79.93 & 88.14 & 67.28 & 84.09          \\
                             & X & 81.79          & 83.68 & 82.65 & 80.88 & 63.32 & 83.13          \\
                             & N & 78.20          & 76.44 & 86.96 & 68.24 & 55.64 & 81.20          \\ 
\midrule
\multirow{2}{*}{DenseNet121} & L & 93.70          & 92.43 & 96.18 & 90.77 & 87.25 & 94.25          \\
                             & S & 92.25          & 89.94 & 96.48 & 87.26 & 84.26 & 93.08          \\                        
\multirow{2}{*}{DenseNet121} 
                             & X & 93.37          & 92.14 & 95.95 & 90.34 & 86.58 & 93.98          \\
                             & N & 88.80          & 83.82 & 98.28 & 77.57 & 77.03 & 90.46          \\ 
\midrule
\multirow{4}{*}{DenseNet169} & L & 84.73          & 88.93 & 82.15 & 87.90 & 69.43 & 85.37          \\
                             & S & 86.41          & 93.87 & 80.25 & 93.77 & 72.96 & 86.51          \\
                             & X & 83.99          & 87.44 & 82.43 & 85.99 & 67.89 & 84.80          \\
                             & N & 79.93          & 79.85 & 84.85 & 74.53 & 59.44 & 82.10          \\ 
\midrule
\multirow{4}{*}{DenseNet201} & L & 93.84          & 93.28 & 95.39 & 92.06 & 87.57 & 94.31          \\
                             & S & 93.42          & 92.12 & 96.04 & 90.38 & 86.68 & 94.02          \\
                             & X & 93.51          & 93.13 & 94.95 & 91.80 & 86.89 & 94.01          \\
                             & N & 91.27          & 88.39 & 96.57 & 85.12 & 82.27 & 92.27          \\ 
\midrule
\multirow{4}{*}{NASNetMobile} & L & 91.41          & 90.00 & 94.64 & 87.64 & 82.60 & 92.24          \\
                             & S & 91.88          & 89.86 & 95.76 & 87.38 & 83.55 & 92.68          \\
                             & X & 90.24          & 89.18 & 93.26 & 86.70 & 80.24 & 91.16          \\
                             & N & 90.71          & 87.11 & 97.24 & 83.11 & 81.10 & 91.86          \\ 
\midrule
\multirow{4}{*}{NASNetLarge} & L & 83.99          & 86.70 & 83.26 & 84.93 & 67.82 & 84.93          \\
                             & S & 86.09          & 92.77 & 80.69 & 92.53 & 72.27 & 86.30          \\
                             & X & 83.99          & 86.19 & 83.95 & 84.12 & 67.79 & 85.03          \\
                             & N & 81.84          & 81.18 & 87.02 & 76.14 & 63.28 & 83.84          \\ 
\bottomrule

\end{longtable}
\end{scriptsize}
\end{clearpage}
}
As evident from the results presented in \autoref{tab:fe_cls_res}, using the Xception model as a feature extractor achieved the highest accuracy among all 52 tested models. Consequently, it was selected as the feature extractor for this study. Boldface values in the table indicate the maximum metrics for the classifiers and feature extractors.

\begin{figure}[htbp]\centering
\includegraphics[width=0.5\linewidth]{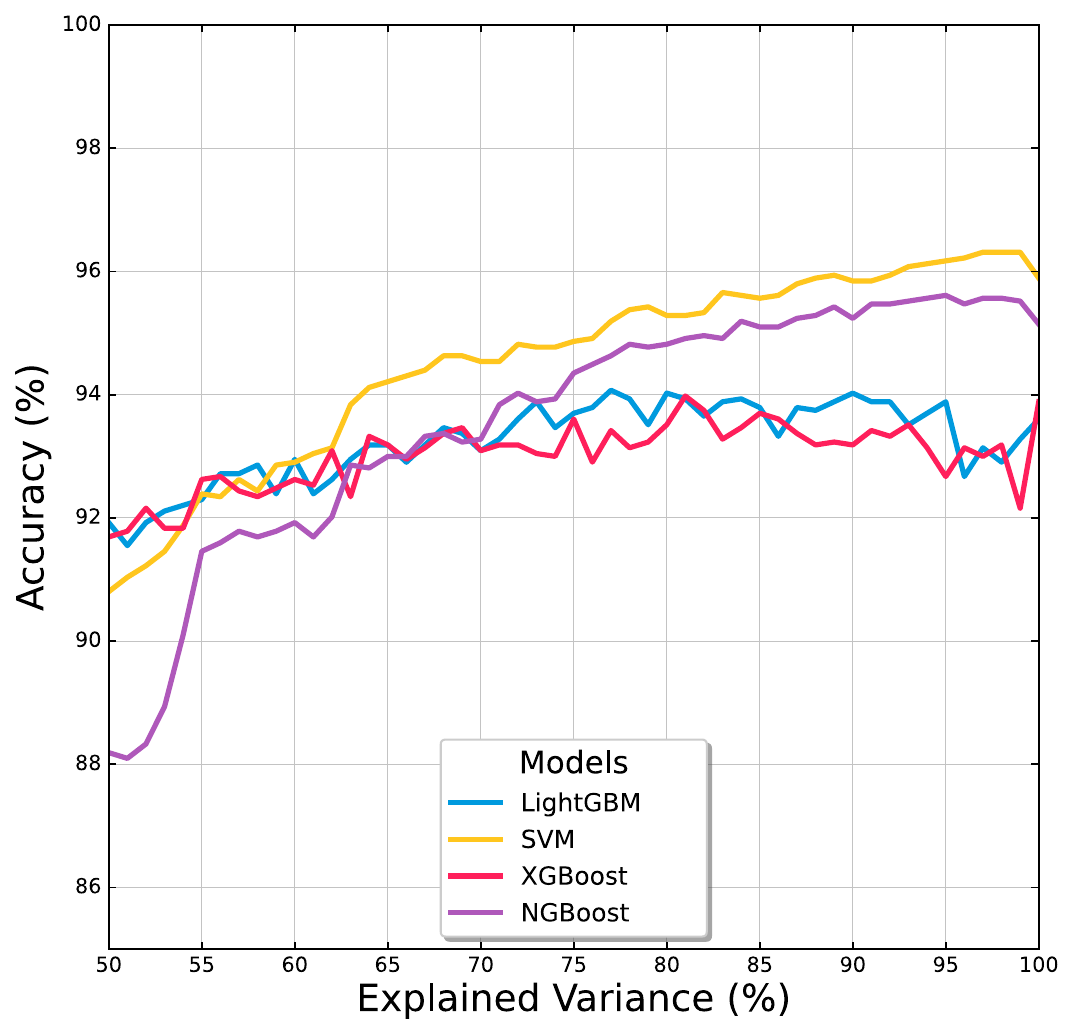}
\caption{Classification performance as a function of PCA variance ratio.}
\label{fig:mls_pca_rat}
\end{figure}

A comprehensive analysis was conducted to determine the optimal variance ratio for PCA. The performance of four classifiers—LightGBM, SVM, XGBoost, and NGBoost—was evaluated across different portions of the variance. The results are depicted in \autoref{fig:mls_pca_rat}, which shows the classification performance scores as a function of the variance ratio.

As shown in \autoref{fig:mls_pca_rat}, the classification performance varied with the variance ratio. Notably, both SVM and NGBoost demonstrated superior performance when the variance ratio was set to 0.95 (266 features) and 0.97 (530 features), respectively. These optimal variance ratios were selected based on the highest classification accuracy achieved.

\autoref{tab:cls_r_with_withoutpca} presents the classification results with and without dimensionality reduction. The table clearly indicates that dimensionality reduction is essential to improve the accuracy of the classification results.

\begin{table}[h]
\caption{Comparison between the classification results with and without dimensionality reduction.}
\label{tab:cls_r_with_withoutpca}
\resizebox{\textwidth}{!}{
\begin{tabular}{@{\extracolsep{\fill}}lcccccccccccc}
\toprule
\multirow{2}{*}{Metric} & \multicolumn{4}{c}{Before Dimensionality Reduction} & \multicolumn{4}{c}{PCA (variance ratio = 0.95)} & \multicolumn{4}{c}{PCA (variance ratio = 0.97)} \\ 
\cmidrule(lr){2-5} \cmidrule(lr){6-9} \cmidrule(lr){10-13}
 & L & S & X & N & L & S & X & N & L & S & X & N \\ 
\midrule
Accuracy (\%)    & 93.60 & 95.89 & 93.51 & 95.15 & 93.00 & 96.17 & 93.60 & 95.61 & 93.18 & 96.31 & 93.42 & 95.57 \\
Precision (\%)   & 92.47 & 94.75 & 92.73 & 93.65 & 91.39 & 95.29 & 92.26 & 94.11 & 92.05 & 95.53 & 92.55 & 94.11 \\
Recall (\%)      & 96.11 & 97.88 & 95.60 & 97.70 & 96.17 & 97.78 & 96.34 & 98.04 & 95.73 & 97.78 & 95.54 & 97.95 \\
Specificity (\%) & 90.79 & 93.63 & 91.20 & 92.20 & 89.35 & 94.35 & 90.53 & 92.83 & 90.26 & 94.67 & 90.99 & 92.82 \\
Kappa            & 0.8705 & 0.9168 & 0.8688 & 0.9016 & 0.8581 & 0.9226 & 0.8706 & 0.9112 & 0.8620 & 0.9255 & 0.8668 & 0.9102 \\
F1-Score (\%)    & 94.22 & 96.27 & 94.10 & 95.61 & 93.70 & 96.50 & 94.21 & 96.02 & 93.83 & 96.62 & 94.01 & 95.67 \\ 
\bottomrule
\end{tabular}%
}
\end{table}

In addition to evaluating classification performance, the impact of PCA on the training times of the four classifiers was analyzed. \autoref{tab:tr_ti} presents the training times for the complete dataset before and after applying PCA.

\begin{table}[h]
\caption{Classifier training times before and after PCA.}
\label{tab:tr_ti}
\resizebox{\textwidth}{!}{
\begin{tabular}{@{\extracolsep{\fill}}lccc}
\toprule
\multirow{2}{*}{Classifier} & \multicolumn{3}{c}{Training Time (sec.)} \\ 
\cmidrule(lr){2-4}
 & Before Dimensionality Reduction & PCA (number of features = 266) & PCA (number of features = 530) \\ 
\midrule
LightGBM & 16.36 & 3.31 & 6.28 \\
SVM      & 3.36  & 0.64 & 0.89 \\
XGBoost  & 28.37 & 7.06 & 9.57 \\
NGBoost  & 85.97 & 17.75 & 22.58 \\ 
\bottomrule
\end{tabular}%
}
\end{table}

\begin{figure}[htbp] 
    \centering
    \begin{subfigure}[t]{0.49\textwidth} 
        \centering
        \includegraphics[width=\linewidth]{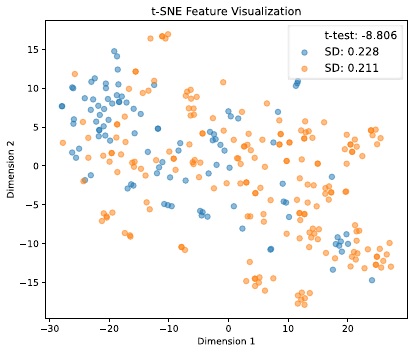} 
        \caption{}
        \label{fig:r2}
    \end{subfigure}
    \hfill
    \begin{subfigure}[t]{0.49\textwidth}
        \centering
        \includegraphics[width=\linewidth]{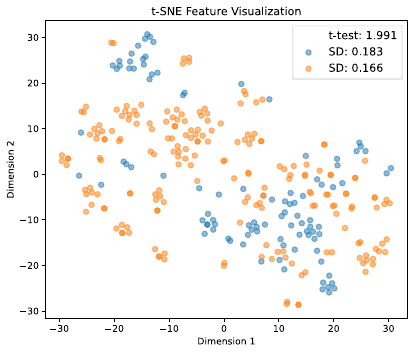}
        \caption{}
        \label{fig:r3}
    \end{subfigure}
    
    \vspace{0.5em} 
    
    \begin{subfigure}[t]{0.49\textwidth}
        \centering
        \includegraphics[width=\linewidth]{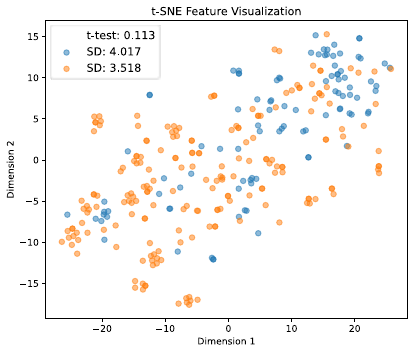}
        \caption{}
        \label{fig:r4}
    \end{subfigure}
    \hfill
    \begin{subfigure}[t]{0.49\textwidth}
        \centering
        \includegraphics[width=\linewidth]{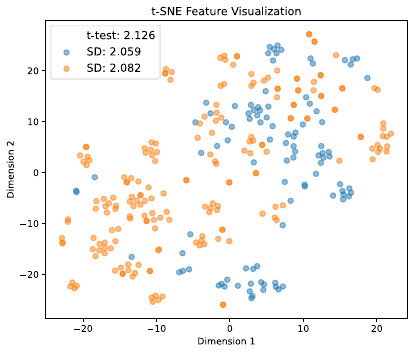}
        \caption{}
        \label{fig:r5}
    \end{subfigure}
    
    \caption{Visualization of feature sets using t-SNE: (a) original data, (b) Xception feature extraction, (c) PCA feature extraction, and (d) combined method of PCA and Xception.}
    \label{fig:r2_5}
\end{figure}

Additionally, the data distribution at various stages was analyzed, as illustrated in \autoref{fig:r2_5}. The original dataset contains 2048 attributes for each individual, representing 2048 dimensions. To visualize the data, the t-distributed stochastic neighbor embedding (t-SNE) method was employed to project this high-dimensional data into a two-dimensional plane. To achieve better clustering visualization for both classes, the experiment was conducted on 300 samples from the dataset, with the Euclidean metric chosen to measure the similarity between the samples.

t-SNE cannot provide class-specific cluster comparison between the Xception network (\autoref{fig:r3}) and the PCA method (\autoref{fig:r4}) because PCA is an unsupervised dimensionality reduction technique. However, a detailed comparison of \autoref{fig:r3} and \autoref{fig:r4} reveals that the points in \autoref{fig:r4} are more concentrated. Comparing \autoref{fig:r3} and \autoref{fig:r5} based on their orange and blue dots, which represent the two classes in the data, demonstrates that the combination of PCA and Xception (\autoref{fig:r5}) results in more distinctly separated groups. Notable differences are also observed in the coordinate axes of \autoref{fig:r3} and \autoref{fig:r5}.

Beyond observing the positions of the blue and orange dots to analyze individual distributions, statistical methods like standard deviation (SD) and t-tests were employed to analyze the values of the two groups. The t-test determines whether the two independent groups exhibit statistically significant differences, with a result approaching zero indicating no difference between the groups. The results are displayed in the upper corner of each panel in \autoref{fig:r2_5}. \autoref{fig:r2} depicts the original data analysis with a t value of -8.806. The SD values for the two categories were 0.228 and 0.211, respectively, serving as the control group. Data processed through the Xception model are presented in \autoref{fig:r3}, where the t value improved to 1.991, and the SD values reduced to 0.183 and 0.166 for the two categories. These improvements were accompanied by observable changes in the curves along the horizontal and vertical axes. The final processed data in \autoref{fig:r5} show a t value of 2.126 and SD values of 2.059 and 2.082 for the two categories, respectively. Although the SD values showed no significant improvement, the t values indicated substantial progress, with the individual distribution groups becoming largely separated. Extracted features from Xception were then transferred to classifiers to establish the final decision boundary, dividing the data into two categories.

Statistical analysis using SD and t-tests highlighted the strengths and weaknesses of the blue and orange dots. The t-test results confirmed that \autoref{fig:r5} produced the most outstanding outcomes. By assessing the two groups using these metrics, a clearer distinction was achieved, validating the superiority of the combined methods.

The AVOA-NGB ensemble invites a good deal of hyperparameter tuning. To address this, we focused on the most sensitive hyperparameters. To set up the model, the Support Vector Regression (SVR) was applied as the base learner. The selection of SVR was due to its superior performance and fewer parameters to be tuned. Only two sensitive parameters—the learning rate and the number of estimators—required optimization. The optimized NGBoost hyperparameters using AVOA and their domain range are detailed in \autoref{tab:opt_ngb_pars}.

\begin{table}[h]
\caption{Optimized NGBoost hyperparameters using AVOA, along with their domain ranges.}
\label{tab:opt_ngb_pars}
\resizebox{\textwidth}{!}{%
\begin{tabular}{@{\extracolsep{\fill}}llll}
\toprule
\multirow{2}{*}{Hyperparameter} & \multirow{2}{*}{Description} & \multicolumn{2}{c}{Values} \\ 
\cmidrule(lr){3-4}
 &  & Domain Range & Optimal Value \\ 
\midrule
learning\_rate & Step size for shrinkage update. & $[1 \times 10^{-7}, \dots, 0.9]$ & $1.0921481 \times 10^{-1}$ \\
n\_estimator & Number of boosting iterations to perform. & $[3, \dots, 20]$ & 5 \\
Base\_learner & Base component of multiple classifier systems. & ('SVR', 'DecisionTreeRegressor', 'Ridge') & SVR \\
Probability\_distribution & Distribution for output type. & ('k\_categorical', 'Bernoulli') & Bernoulli \\ 
\bottomrule
\end{tabular}%
}
\end{table}

This paper uses AVOA to optimize the hyperparameter of four classiﬁcation algorithms. \autoref{tab:avo_pars} displays the proposed algorithm's conﬁguration parameters. These parameters include the population size, number of training iterations, \textit{P1, P2, P3} and ﬁnally, the values of alpha and gamma, which were set as 0.8 and 2.5, respectively.

\begin{table}[h]
\centering
\caption{The proposed controlling parameter values.}
\label{tab:avo_pars}
\footnotesize
\begin{tabular}{ll}
\toprule
Parameter & Value \\ 
\midrule
Population size    & 50             \\
Number of iterations & 40           \\
$\gamma$           & 2.5            \\
$\alpha$           & 0.8            \\
$P_1$              & 0.6            \\
$P_2$              & 0.4            \\
$P_3$              & 0.6            \\
\bottomrule
\end{tabular}
\end{table}

As described, the NGBoost classifier was trained on 80\% of the dataset and tested on the remaining 20\%. The training phase involved tuning the parameters of the SVR base learner and optimizing the NGBoost parameters.

To evaluate the effectiveness of the proposed technique, accuracy and computational time were assessed. Accuracy comparisons were performed across three algorithm variations: (1) the baseline NGBoost algorithm, (2) NGBoost with reduced features but without hyperparameter optimization, and (3) AVOA-NGB with reduced features and optimized \textit{l\_rate} and \textit{n\_est}.

As seen in \autoref{tab:acc_comtime}, the proposed AVOA-NGB method with dimensionality reduction and optimized hyperparameters provided the highest accuracy. Regarding computational time, the baseline algorithm was the slowest due to processing the entire feature set. In contrast, the AVOA-NGB method employed PCA-based dimensionality reduction, eliminating approximately 75\% of redundant features and accelerating computation by nearly an order of magnitude. This significant improvement in computational efficiency indicates the feasibility of implementing the method in real-time applications.

\begin{table}[h]
\caption{Accuracy and computational time comparison of three algorithm variations.}
\label{tab:acc_comtime}
\resizebox{\textwidth}{!}{
\begin{tabular}{@{\extracolsep{\fill}}lccc}
\toprule
\multirow{2}{*}{Algorithm} & \multirow{2}{*}{Number of Features} & \multicolumn{2}{c}{Performance Metrics} \\ 
\cmidrule(lr){3-4}
 &  & Accuracy (\%) & Computational Time (sec.) \\ 
\midrule
Baseline NGBoost & Whole feature set (2048) & 95.15 & 0.0054 \\
NGBoost with PCA & 530 & 95.57 & 0.0012 \\
NGBoost with PCA and optimized NGBoost hyperparameters & 530 & 97.53 & 0.0003 \\ 
\bottomrule
\end{tabular}%
}
\end{table}

To demonstrate the performance of the tuning strategies, the AVOA is compared with other meta-heuristic optimizers, namely the Harris hawks optimization (HHO) \citep{Heidari2019}, hybrid improved whale optimization algorithm (HI-WOA) \citep{Tang2019}, modified gorilla troops optimization (mGTO) \citep{You2023}, hybrid grey wolf - whale optimization algorithm (GWO-WOA) \citep{Dhakhinamoorthy2023} and Q-learning embedded sine cosine algorithm (QLESCA) \citep{Hamad2022}. For a fair comparison, all competing optimizers utilized the same set of candidate inputs, preprocessing methods, training and validation samples, and forecast samples. The only variation was the optimizer type (e.g., replacing AVOA-NGB with HHO-NGB, etc.). Based on the tests, the AVOA-NGB exhibited the best performance. Thus, it can be concluded that the AVOA has superior optimization capability compared to other methods. \autoref{tab:competing_alg_pars} displays the settings for the other competing algorithms used in the experiments. 

\begin{table}[h]
\caption{Configuration parameters of the competing algorithms.}
\label{tab:competing_alg_pars}
\begin{tabularx}{\textwidth}{@{}lXl@{}}
\toprule
Algorithm & Parameter & Value \\ 
\midrule
\multirow{2}{*}{HHO} & Population size & 50 \\
 & Number of iterations & 40 \\ 
\hline
\multirow{3}{*}{HIWOA} & Population size & 50 \\
 & Number of iterations & 40 \\
 & Maximum iterations of each feedback & 10 \\ 
\hline
\multirow{3}{*}{mGTO} & Population size & 50 \\
 & Number of iterations & 40 \\
 & Probability of transition in exploration phase ($P$) & 0.03 \\ 
\hline
\multirow{2}{*}{GWO\_WOA} & Population size & 50 \\
 & Number of iterations & 40 \\ 
\hline
\multirow{4}{*}{QLESCA} & Population size & 50 \\
 & Number of iterations & 40 \\
 & Learning rate in Q-learning ($\alpha$) & 0.1 \\
 & Discount factor ($\gamma$) & 0.9 \\ 
\bottomrule
\end{tabularx}
\end{table}

\begin{figure}[htbp]\centering
\includegraphics[width=0.5\linewidth]{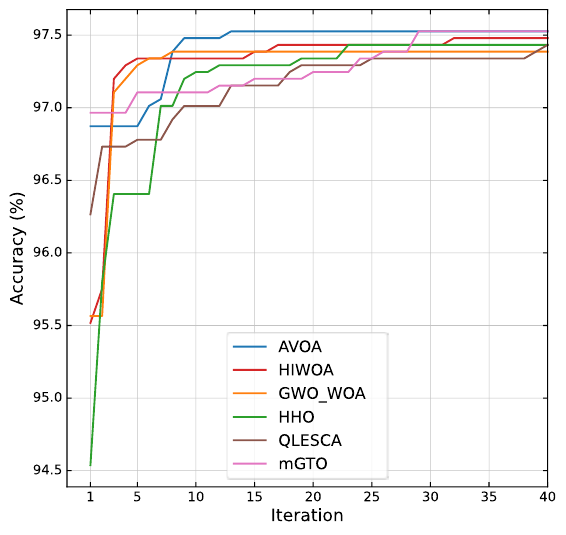}
\caption{Convergence curves of average fitness function for parameter optimization methods.}
\label{fig:conv_curve}
\end{figure}

\autoref{fig:conv_curve} illustrates the convergence curves of the average fitness function for various parameter optimization methods. The ordinate represents the maximum accuracy achieved, while the abscissa denotes the number of iterations. The figure shows that GWO-WOA and QLESCA converge more slowly. HI-WOA converges faster and outperforms GWO-WOA and QLESCA. Although mGTO achieves slightly higher accuracy than HI-WOA, HI-WOA reaches convergence more quickly. AVOA and mGTO demonstrate the fastest convergence among the optimization methods. GWO-WOA reached the convergence state early, which resulted in its unsatisfactory final accuracy and failure to find the global optimal value. AVOA achieved the best performance among all the convergence performances. Furthermore, mGTO continued to improve after several iterations, indicating that its optimization mechanism promotes subgroup diversity, prevents local particle clustering, and enhances the search for local optima.

The performance of AVOA with NGBoost was compared with other classifiers: LightGBM, SVM, and XGBoost. Each of these models featured parameters influencing their classification performance. Hence, the AVOA was employed to find the optimal values for the classifiers' parameters. The optimized values of the hyperparameters using AVOA and their domain ranges are shown in \autoref{tab:hy_mls}.

\begin{table}[h]
\caption{Optimized hyperparameters and their domain ranges for classifiers using AVOA}
\label{tab:hy_mls}
\footnotesize
\begin{tabular*}{\textwidth}{@{\extracolsep{\fill}} l l l l}
\toprule
Classifier & Hyper-parameters & Domain Range & Optimal Value \\ 
\midrule
\multirow{5}{*}{LightGBM} & Number of decision trees & $[15, \dots, 200]$ & 84 \\ 
                          & Maximum depth of a tree & $[5, \dots, 300]$ & 127 \\ 
                          & Number of boosting iterations & $[70, \dots, 300]$ & 100 \\ 
                          & Learning rate & $[1 \times 10^{-7}, \dots, 0.9]$ & $2.405572 \times 10^{-1}$ \\ 
                          & Maximum number of leaves on one tree & $[5, \dots, 400]$ & 291 \\
\midrule
\multirow{4}{*}{SVM}      & Kernel & ('linear', 'rbf') & rbf \\
                          & Cost & $[0.1, \dots, 20]$ & 14.681521 \\
                          & Kernel coefficient ($\gamma$) & $[1 \times 10^{-4}, \dots, 0.9]$ & $6.535413 \times 10^{-3}$ \\
                          & Probability estimates & - & True \\
\midrule
\multirow{5}{*}{XGBoost}  & Base Learner & ('gblinear', 'gbtree') & Gradient boosted tree \\
                          & Learning rate ($\eta$) & $[1 \times 10^{-7}, \dots, 0.9]$ & $2.863612 \times 10^{-2}$ \\
                          & Lagrange multiplier ($\gamma$) & $[1 \times 10^{-5}, \dots, 0.9]$ & $1.571300 \times 10^{-3}$ \\
                          & Maximum depth of a tree & $[5, \dots, 200]$ & 23 \\
                          & Maximum number of leaves on one tree & $[5, \dots, 800]$ & 720 \\
\bottomrule
\end{tabular*}
\end{table}

As illustrated in \autoref{tab:metrc_all_mls}, the proposed method demonstrates superior performance across all classification metrics, achieving accuracy (97.53\%), Kappa value (0.95), F1-Score (97.72\%), and AUC (97.47\%). The AVOA-SVM method ranked second. In contrast, the accuracy of LightGBM with PCA (93.18\%) was the lowest among the classifiers examined in this study. To highlight the significance of AVOA, results from models without AVOA on the same input data are presented. It is clear that all classifiers showed reduced performance when AVOA was not utilized. For instance, the accuracy (95.89\%), Cohen Kappa (0.9168), and F1-Score (96.27\%) of SVM without AVOA were notably lower than those of the AVOA-optimized SVM.

\begin{table}[h]
\caption{Classification metrics for various models.}
\label{tab:metrc_all_mls}
\resizebox{\textwidth}{!}{%
\begin{tabular}{@{\extracolsep{\fill}}lccccccc}
\toprule
\multirow{2}{*}{Classifiers} & \multicolumn{7}{c}{Performance Metrics} \\ 
\cmidrule(lr){2-8}
 & Accuracy (\%) & Precision (\%) & Recall (\%) & Specificity (\%) & Kappa & F1-Score (\%) & AUC (\%) \\ 
\midrule
LightGBM                & 93.60 & 92.47 & 96.11 & 90.79 & 0.8705 & 94.22 & 93.45 \\
SVM                     & 95.89 & 94.75 & 97.88 & 93.63 & 0.9168 & 96.27 & 95.75 \\
XGBoost                 & 93.51 & 92.73 & 95.60 & 91.20 & 0.8688 & 94.10 & 93.75 \\
NGBoost                 & 95.15 & 93.65 & 97.70 & 92.20 & 0.9016 & 95.61 & 94.95 \\ 
AVOA-LGBM               & 94.12 & 93.67 & 95.68 & 92.40 & 0.8811 & 94.64 & 94.04 \\
AVOA-SVM                & 96.87 & 96.42 & 97.90 & 95.78 & 0.9368 & 97.13 & 96.84 \\
AVOA-XGB                & 94.44 & 93.65 & 96.38 & 92.30 & 0.8876 & 94.96 & 94.12 \\
AVOA-NGB                & 97.20 & 97.08 & 97.80 & 96.60 & 0.9435 & 97.42 & 97.20 \\
LightGBM + PCA          & 93.18 & 92.05 & 95.73 & 90.26 & 0.8620 & 93.83 & 92.92 \\
SVM + PCA               & 96.31 & 95.53 & 97.78 & 94.67 & 0.9255 & 96.62 & 96.23 \\
XGBoost + PCA           & 93.42 & 92.55 & 95.54 & 90.99 & 0.8668 & 94.01 & 92.86 \\
NGBoost + PCA           & 95.57 & 94.11 & 97.95 & 92.82 & 0.9102 & 95.97 & 95.39 \\
AVOA-LGBM + PCA         & 93.88 & 92.62 & 96.41 & 91.04 & 0.8763 & 94.45 & 92.88 \\
AVOA-SVM + PCA          & 96.78 & 96.18 & 97.98 & 95.49 & 0.9349 & 97.05 & 96.67 \\
AVOA-XGB + PCA          & 94.17 & 93.11 & 96.43 & 91.67 & 0.8821 & 94.69 & 94.01 \\
Proposed Method         & \textbf{97.53} & \textbf{97.34} & \textbf{98.15} & \textbf{96.90} & \textbf{0.9500} & \textbf{97.72} & \textbf{97.47} \\
\bottomrule
\end{tabular}%
}
\end{table}

The results obtained by the model for different folds, along with the standard deviation and the mean of all folds, are presented in \autoref{tab:cross_fo}.

\begin{table}[h]
\caption{5-fold cross-validation results.}
\label{tab:cross_fo}
\resizebox{\textwidth}{!}{
\begin{tabular}{@{\extracolsep{\fill}}lcccccc}
\toprule
\multirow{2}{*}{Number of Fold} & \multicolumn{6}{c}{Evaluation Metrics} \\ 
\cmidrule(lr){2-7}
 & Accuracy (\%) & Precision (\%) & Recall (\%) & Specificity (\%) & Kappa & F1-Score (\%) \\ 
\midrule
Fold 1 & 96.503 & 96.34 & 97.53 & 95.16 & 0.9287 & 96.93 \\
Fold 2 & 96.736 & 94.20 & 99.53 & 94.01 & 0.9384 & 96.79 \\
Fold 3 & 98.364 & 98.73 & 98.32 & 98.42 & 0.9669 & 98.53 \\
Fold 4 & 97.663 & 99.14 & 96.65 & 98.94 & 0.9528 & 97.88 \\
Fold 5 & 98.364 & 98.27 & 98.70 & 97.98 & 0.9671 & 98.48 \\ 
\midrule
\textbf{Mean} & \textbf{97.53} & \textbf{97.34} & \textbf{98.15} & \textbf{96.90} & \textbf{0.9500} & \textbf{97.72} \\
\textbf{Standard Deviation} & \textbf{0.78} & \textbf{1.84} & \textbf{0.99} & \textbf{1.95} & \textbf{0.015} & \textbf{0.74} \\
\bottomrule
\end{tabular}%
}
\end{table}

The analysis reveals a low computed standard deviation, indicating consistent performance by the proposed model across all folds, with outstanding results observed in folds 3 and 5.  The low variation in scores across folds suggests that the model is robust and does not exhibit high variance or overfitting on any single fold. The model's strong performance on multiple distinct subsets of the data, as represented by the different folds, supports its generalizability and reliability as an accurate predictive model for this problem.

\begin{figure}[htbp]
    \centering
    \begin{subfigure}[t]{0.32\textwidth} 
        \centering
        \includegraphics[width=\linewidth]{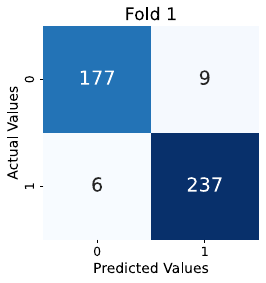}
        \label{fig:cm1}
    \end{subfigure}
    \hfill
    \begin{subfigure}[t]{0.32\textwidth}
        \centering
        \includegraphics[width=\linewidth]{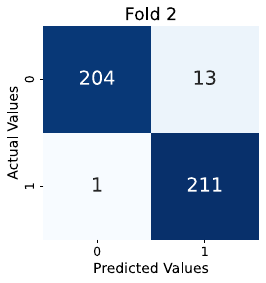}
        \label{fig:cm2}
    \end{subfigure}
    \hfill
    \begin{subfigure}[t]{0.32\textwidth}
        \centering
        \includegraphics[width=\linewidth]{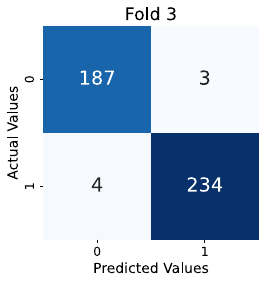}
        \label{fig:cm3}
    \end{subfigure}
    
    \vspace{1em} 
    
    \begin{subfigure}[t]{0.32\textwidth}
        \centering
        \includegraphics[width=\linewidth]{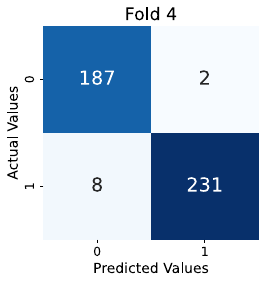}
        \label{fig:cm4}
    \end{subfigure}
    \hfill
    \begin{subfigure}[t]{0.32\textwidth}
        \centering
        \includegraphics[width=\linewidth]{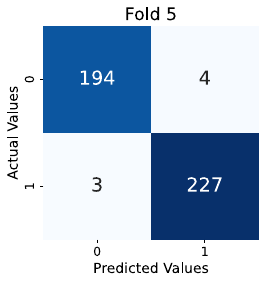}
        \label{fig:cm5}
    \end{subfigure}
    \hfill
    \begin{subfigure}[t]{0.32\textwidth}
        \centering
        \includegraphics[width=\linewidth]{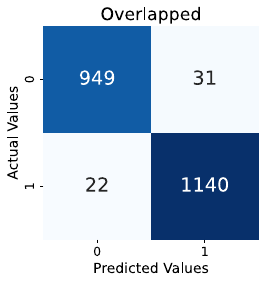}
        \label{fig:cm_ov}
    \end{subfigure}
    
    \caption{Visualization of confusion matrices for proposed model.}
    \label{fig:conf_matrices}
\end{figure}

To examine the model's patterns, a confusion matrix was computed for each fold and then overlapped, as depicted in \autoref{fig:conf_matrices}. The combined entries from all folds form the overlapping confusion matrix, which indicates that the proposed architecture correctly classified monkeypox cases with 98.10\% accuracy and non-monkeypox cases with 96.84\% accuracy. These results highlight the method's effectiveness in detecting true positive samples. Furthermore, the first fold, demonstrated the weakest performance, with 15 misclassifications out of 429 test samples.

\begin{figure}[htbp]\centering
\includegraphics[width=0.5\linewidth]{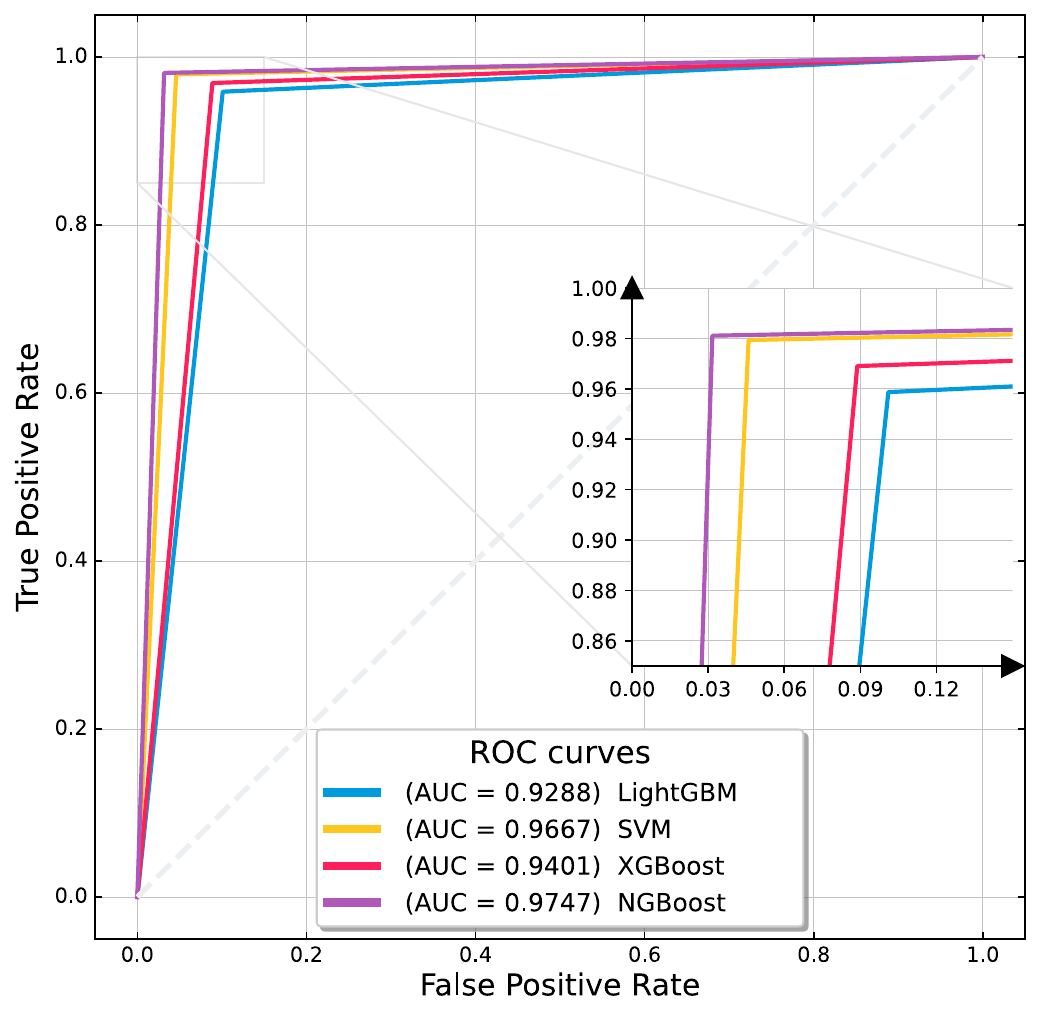}
\caption{ROC curves of different classifiers.}
\label{fig:roc_curve}
\end{figure}

The receiver operating characteristic (ROC) curve and its area under the curve (AUC) are employed to compare our model with previous state-of-the-art prediction models. An ROC curve visualizes the true positive rate against the false positive rate under varying thresholds. This enables the evaluation and selection of classification models based on user-specific requirements, typically linked to variable error costs and efficiency assumptions. The plot illustrates the trade-off between recall and specificity, while the AUC quantifies the model's predictive ability. A higher AUC indicates greater usefulness, demonstrating the classifier's capability to differentiate between various classes. ROC curves that bend sharply toward the upper-left corner indicate better performance, reflecting high recall with a low false positive rate. \autoref{fig:roc_curve} illustrates these results, representing the AUC in a magnified format. The ROC curve reveals that our proposed model outperformed other classifiers.

The decisions made by deep learning models can be elucidated using eXplainable Artificial Intelligence (XAI) techniques. Two methods used in this study are Locally Interpretable Model-Agnostic Explanations (LIME) and Grad-CAM.

\begin{figure}[htbp]
    \centering
    \begin{subfigure}[t]{0.19\textwidth} 
        \centering
        \includegraphics[width=3cm,height=3cm]{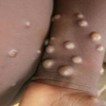}
        \caption{}
        \label{fig:o1}
    \end{subfigure}
    \hfill
    \begin{subfigure}[t]{0.19\textwidth}
        \centering
        \includegraphics[width=3cm,height=3cm]{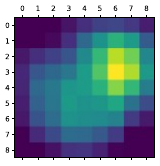}
        \caption{}
        \label{fig:hm1}
    \end{subfigure}
    \hfill
    \begin{subfigure}[t]{0.19\textwidth}
        \centering
        \includegraphics[width=3cm,height=3cm]{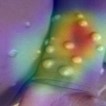}
        \caption{}
        \label{fig:gc1}
    \end{subfigure}
    \hfill
    \begin{subfigure}[t]{0.19\textwidth}
        \centering
        \includegraphics[width=3cm,height=3cm]{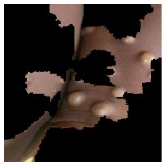}
        \caption{}
        \label{fig:li1}
    \end{subfigure}
    \hfill
    \begin{subfigure}[t]{0.19\textwidth}
        \centering
        \includegraphics[width=3.5cm,height=3.5cm]{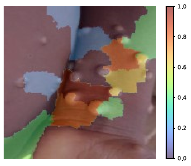}
        \caption{}
        \label{fig:sup1}
    \end{subfigure}
    
    \vspace{1em} 

    \begin{subfigure}[t]{0.19\textwidth}
        \centering
        \includegraphics[width=3cm,height=3cm]{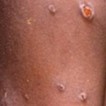}
        \caption{}
        \label{fig:o2}
    \end{subfigure}
    \hfill
    \begin{subfigure}[t]{0.19\textwidth}
        \centering
        \includegraphics[width=3cm,height=3cm]{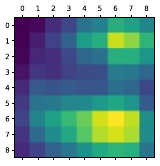}
        \caption{}
        \label{fig:hm2}
    \end{subfigure}
    \hfill
    \begin{subfigure}[t]{0.19\textwidth}
        \centering
        \includegraphics[width=3cm,height=3cm]{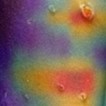}
        \caption{}
        \label{fig:gc2}
    \end{subfigure}
    \hfill
    \begin{subfigure}[t]{0.19\textwidth}
        \centering
        \includegraphics[width=3cm,height=3cm]{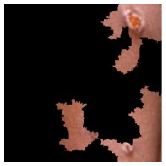}
        \caption{}
        \label{fig:li2}
    \end{subfigure}
    \hfill
    \begin{subfigure}[t]{0.19\textwidth}
        \centering
        \includegraphics[width=3.5cm,height=3.5cm]{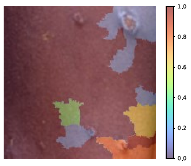}
        \caption{}
        \label{fig:sup2}
    \end{subfigure}
    
    \vspace{4em} 

    \begin{subfigure}[t]{0.19\textwidth}
        \centering
        \includegraphics[width=3cm,height=3cm]{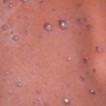}
        \caption{}
        \label{fig:o3}
    \end{subfigure}
    \hfill
    \begin{subfigure}[t]{0.19\textwidth}
        \centering
        \includegraphics[width=3cm,height=3cm]{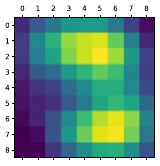}
        \caption{}
        \label{fig:hm3}
    \end{subfigure}
    \hfill
    \begin{subfigure}[t]{0.19\textwidth}
        \centering
        \includegraphics[width=3cm,height=3cm]{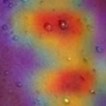}
        \caption{}
        \label{fig:gc3}
    \end{subfigure}
    \hfill
    \begin{subfigure}[t]{0.19\textwidth}
        \centering
        \includegraphics[width=3cm,height=3cm]{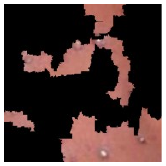}
        \caption{}
        \label{fig:li3}
    \end{subfigure}
    \hfill
    \begin{subfigure}[t]{0.19\textwidth}
        \centering
        \includegraphics[width=3.5cm,height=3.5cm]{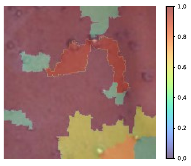}
        \caption{}
        \label{fig:sup3}
    \end{subfigure}
    
    \vspace{1em} 

    \begin{subfigure}[t]{0.19\textwidth}
        \centering
        \includegraphics[width=3cm,height=3cm]{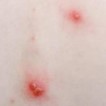}
        \caption{}
        \label{fig:o4}
    \end{subfigure}
    \hfill
    \begin{subfigure}[t]{0.19\textwidth}
        \centering
        \includegraphics[width=3cm,height=3cm]{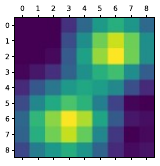}
        \caption{}
        \label{fig:hm4}
    \end{subfigure}
    \hfill
    \begin{subfigure}[t]{0.19\textwidth}
        \centering
        \includegraphics[width=3cm,height=3cm]{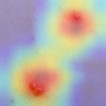}
        \caption{}
        \label{fig:gc4}
    \end{subfigure}
    \hfill
    \begin{subfigure}[t]{0.19\textwidth}
        \centering
        \includegraphics[width=3cm,height=3cm]{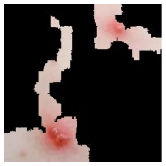}
        \caption{}
        \label{fig:li4}
    \end{subfigure}
    \hfill
    \begin{subfigure}[t]{0.19\textwidth}
        \centering
        \includegraphics[width=3.5cm,height=3.5cm]{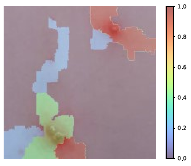}
        \caption{}
        \label{fig:sup4}
    \end{subfigure}
    
    \caption{(a) and (f) represent the class 'Monkeypox,' and (k) and (p) represent the class 'Normal.' (b) and (g) show the Heat-maps for class 'Monkeypox,' while (l) and (q) show the Heat-maps for class 'Normal.' (c) and (h) provide the Grad-CAM visualizations for class 'Monkeypox,' and (m) and (r) for class 'Normal.' (d), (i), (n), and (s) display the top 7 features from the LIME visualizations, and (e), (j), (o), and (t) illustrate their respective LIME images.}
    \label{fig:gr_lime}
\end{figure}

LIME is a visualization tool that highlights important features identified by the convolutional neural network during training to make predictions. It explains the classifier's predictions by locally approximating them using an interpretable model \citep{Nayak2023}. A LIME object is generated with a specific query point and a defined number of significant predictors. This process produces a synthetic dataset, which is then used to train a simple, comprehensible model that explains the predictions for the synthetic data centered around the query point. LIME is illustrated in \autoref{fig:sup1} and \autoref{fig:sup2} for the 'Monkeypox' class and \autoref{fig:sup3} and \autoref{fig:sup4} for the 'Normal' class. The superpixels in these figures are generated using a regression tree, with warmer regions indicating a higher importance of features that are more responsible for making predictions and cooler areas signifying lesser importance in the decision-making process.

These findings are supported by Grad-CAM, as shown in \autoref{fig:gc1} and \autoref{fig:gc2} for the 'Monkeypox' class and \autoref{fig:gc3} and \autoref{fig:gc4} for the 'Normal' class. Grad-CAM is a visualization technique designed to illustrate the "thinking" process of a CNN, enabling users to understand the model's decision-making more effectively. It identifies the regions in an input image that are most influential for a specific class, using gradient information from the final convolutional layers. By assigning significance values to each neuron and generating a heatmap, Grad-CAM highlights the areas the model deems critical for classification. This approach not only provides valuable insights into the internal logic of CNNs but also supports clinical applications by emphasizing regions relevant to diagnosis. In our particular scenario, Grad-CAM focuses primarily on the contaminated areas of the various disease classes, where the primary signs of sickness can be discovered.

\autoref{fig:gr_lime} demonstrates that the Xception model effectively identifies discriminating regions for classification. Grad-CAM highlights virus-infected areas in yellow or dark yellow, while LIME is able to encircle the potentially infected regions with its super pixel on the map. For Grad-CAM, the 'block14 sepconv2 act' layer of the Xception model was used. For LIME, parameters were set as follows: the number of features to 5, the number of samples to 1,000, and the top labels to 5 for the Xception model. Together, these techniques showcase the model's ability to learn and localize infected regions accurately.

\section{Conclusion}
\label{sec:conc}
This work marks a significant advancement in deep learning applications for infectious disease detection, particularly in under-resourced settings. Future studies could explore real-time diagnostic tools integrated into mobile healthcare systems, enabling prompt diagnostic support in remote areas.

\bibliographystyle{plainnat}
\bibliography{library}

\end{document}